\documentclass[11pt, a4paper, logo]{gdmstyle/googledeepmind}
\pdftrailerid{redacted}

\makeatletter
\renewcommand\bibentry[1]{\nocite{#1}{\frenchspacing\@nameuse{BR@r@#1\@extra@b@citeb}}}
\makeatother

\usepackage{algorithm}
\usepackage{algpseudocode}
\newcommand{\algrule}[1][.2pt]

\algnewcommand\algorithmicinput{\textbf{Input:}}
\algnewcommand\Input{\item[\algorithmicinput]}
\algnewcommand\algorithmicoutput{\textbf{Output:}}
\algnewcommand\Output{\item[\algorithmicoutput]}
\algnewcommand\algorithmicempty{~}
\algnewcommand\Empty{\item[\algorithmicempty]}

\usepackage{mathtools}

\usepackage{dsfont}
\usepackage{gdmstyle/gdm-colors}
\usepackage{bm}      %

\usepackage{ifthen}
\newboolean{isanon}
\setboolean{isanon}{false}

\newboolean{islong}
\setboolean{islong}{true}
\usepackage{xcolor}
\usepackage{xurl}
\usepackage{graphicx}
\usepackage{amsmath}

\usepackage{pifont}%
\usepackage[sort,numbers,square]{natbib}
\usepackage[pagebackref=true,colorlinks=true,allcolors=colorgreenfull]{hyperref}
\usepackage[nameinlink]{cleveref}
\usepackage{scalerel}
\usepackage{caption}
\usepackage{wrapfig}
\usepackage{subcaption}
\usepackage{bbding}
\usepackage{placeins}
\usepackage{microtype}
\usepackage{csquotes}
\usepackage{enumitem}
\usepackage{xspace}
\usepackage{booktabs}
\usepackage{mathtools}
\usepackage{amsthm}

\usepackage{tikz}
\usetikzlibrary{calc}
\usetikzlibrary{angles,quotes} %

\usepackage[super]{nth}
\usepackage{multirow}
\usepackage{rotating}
\usepackage{amssymb}
\usepackage[export]{adjustbox}
\usepackage{subcaption}
\usepackage{tcolorbox}
\everypar{\looseness=-1}

\newtheorem{theorem}{Theorem}
\newtheorem*{theorem*}{Theorem}
\Crefname{theorem}{Theorem}{Theorems}

\newtheorem*{assumption*}{Assumption}
\Crefname{assumption}{Assumption}{Assumptions}
\newtheorem{lemma}{Lemma}
\newtheorem*{lemma*}{Lemma}
\Crefname{lemma}{Lemma}{Lemmas}

\Crefname{remark}{Remark}{Remarks}
\Crefname{proposition}{Proposition}{Propositions}
\let\cref\Cref

\usepackage{thmtools} %
\usepackage{longtable}
\usepackage{makecell}
\usepackage{shortcuts}

\newcommand{\KL}{\op{KL}}
\newcommand{\rvec}{\bm{R}}
\newcommand{\mixmap}{f_{\textsf{mix}}}
\newcommand{\piref}{\pi_{\text{ref}}}
\newcommand{\thetaref}{\theta_{\text{ref}}}
\newcommand{\zetaref}{\zeta_{\text{ref}}}
\newcommand{\alphamin}{\alpha_{\min}}

\newcommand{\clp}{{\small\textsf{CLP}}\xspace}
\newcommand{\clpall}{{\small\textsf{full-CLP}}\xspace}
\newcommand{\clpa}{{\small\textsf{attn-CLP}}\xspace}
\newcommand{\clpl}{{\small\textsf{logit-CLP}}\xspace}
\newcommand{\clpp}{{\small\textsf{prompting}}\xspace}

\newcommand{\RRouge}{R_{\text{rouge}}}
\newcommand{\RNLI}{R_{\text{nli}}}
\newcommand{\RTLDR}{R_{\text{tldr}}}
\newcommand{\CondPrompt}{\textsc{CondPrompt}}

\usepackage{makecell}
\usepackage{multirow, tabularx}
\definecolor{seagreen}{rgb}{0.18, 0.55, 0.34}
\definecolor{brickred}{rgb}{0.8, 0.25, 0.33}

\title{Conditional Language Policy: A General Framework for Steerable Multi-Objective Finetuning}

\newcommand{\addLink}[1]{{\normalfont\href{mailto:#1}{\nolinkurl{#1}}}}
\correspondingauthor{Kaiwen Wang (\addLink{wangkaiwen998@gmail.com}), Rahul Kidambi (\addLink{rahulkidambi@google.com}), Alekh Agarwal (\addLink{alekhagarwal@google.com}), Edouard Leurent (\addLink{edouardl@google.com}). Kaiwen Wang and Ryan Sullivan completed this work at Google as student researchers.}

\author{Kaiwen Wang}
\author{Rahul Kidambi}
\author{Ryan Sullivan}
\author{Alekh Agarwal}
\author{Christoph Dann}
\author{Andrea Michi}
\author{Marco Gelmi}
\author{Yunxuan Li}
\author{Raghav Gupta}
\author{Avinava Dubey}
\author{Alexandre Ram\'e}
\author{Johan Ferret}
\author{Geoffrey Cideron}
\author{Le Hou}
\author{Hongkun Yu}
\author{Amr Ahmed}
\author{Aranyak Mehta}
\author{L\'eonard Hussenot}
\author{Olivier Bachem}
\author{Edouard Leurent}
\affil{Google}

\date{\today}

\begin{abstract}
Reward-based finetuning is crucial for aligning language policies with intended behaviors (\emph{e.g.}, creativity and safety). A key challenge is to develop steerable language models that trade-off multiple (conflicting) objectives in a flexible and efficient manner. This paper presents Conditional Language Policy (\clp), a general framework for finetuning language models on multiple objectives. Building on techniques from multi-task training and parameter-efficient finetuning, \clp learn steerable models that effectively trade-off conflicting objectives at {\em inference time}. Notably, this does not require training or maintaining multiple models to achieve different trade-offs between the objectives. Through extensive experiments and ablations on two summarization datasets, we show that \clp learns steerable language models that outperform and Pareto-dominate the existing approaches for multi-objective finetuning.
\end{abstract}

\begin{document}

\maketitle

\section{Introduction}
Reinforcement Learning (RL) is a crucial step for finetuning language models (LMs) to produce desirable, human-aligned behaviors \citep{christiano2017deep,wu2018learning} in various domains such as summarization \citep{ziegler2019fine,stiennon2020learning}, conversational agents \citep{ouyang2022training} and encoding social norms \citep{bai2022constitutional}.
In practice, defining the scalar reward function for RL is challenging due to the plurality of human preferences (\eg, succinctness vs. completeness, factuality vs. creativity) and of applications (\eg, summarization, coding, dialog) \citep{sorensen2024position}. The common practice is to linearly combine multiple, often conflicting rewards with \emph{weightings} that represent the relative importance of each reward \citep{bai2022constitutional,achiam2023gpt}. However, this is not ideal since the model quality is sensitive to the choice of weightings, and more importantly, the model still fails to capture pluralistic alignment since the weightings are fixed.

\begin{figure}[t!]
  \centering
  \includegraphics[width=\linewidth]{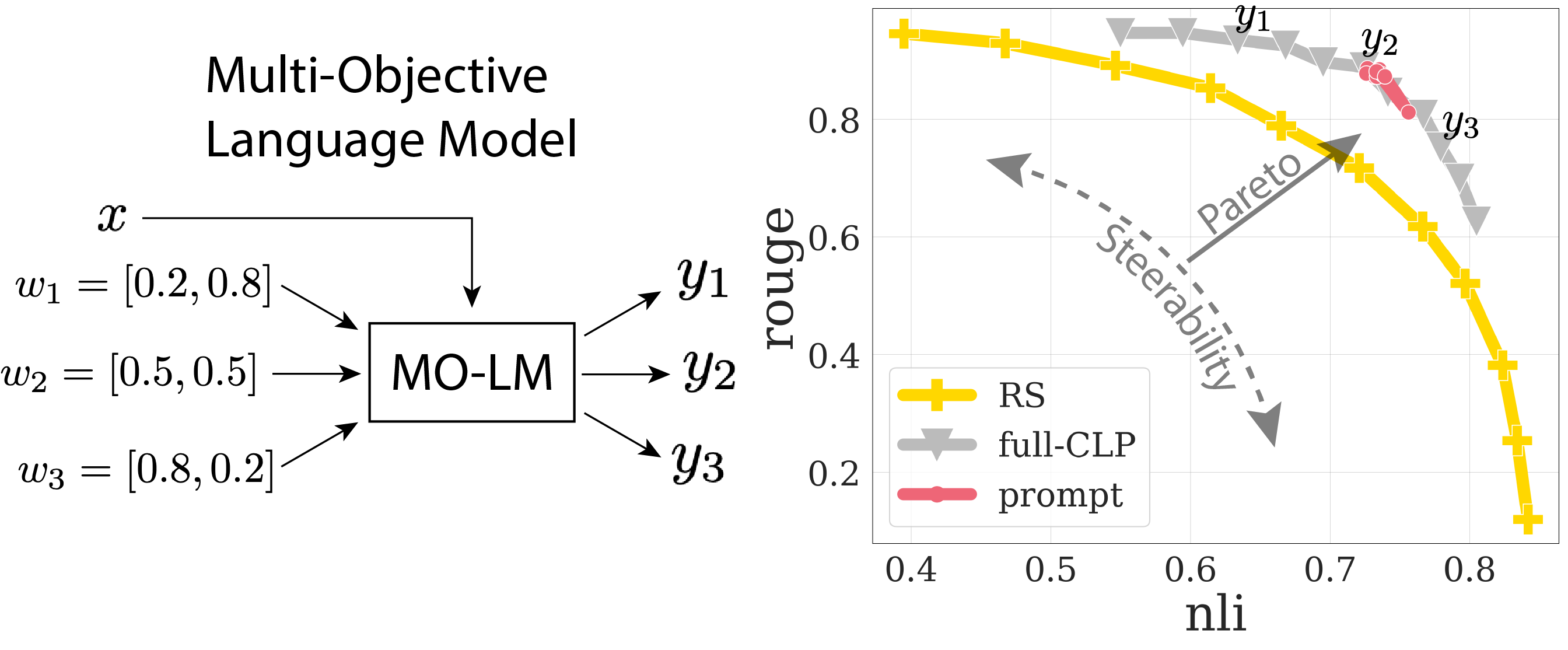}
  \caption{(Left)
  For a fixed prompt $x$, a multi-objective LM can output $y_1,y_2,y_3$ for different weightings $w_1,w_2,w_3$ of two rewards $r_1$ and $r_2$, such that the response $y_i$ for weighting $w_i$ is preferred under the weighted reward $w_i[1] r_1 + w_i[2] r_2$.
  (Right) Pareto-fronts when using the rewards NLI and Rouge (\cref{sec:two-rewards-fixed-kl-regularizer}). Rewarded Soups (RS)~\citep{rame2023soups} is \emph{Pareto-dominated} by both \clpall ({\em this paper}) and \clpp (say,~\citet{jang2023personalized}), but \clpall is more appealing for its steerability, evidenced by its wider Pareto-front. In sum, Pareto-dominance (pushing out the front) and steerability (stretching out the front) are both key desiderata for MOFT.
  }
  \label{fig:figure1}
\end{figure}

A solution we explore in this paper is to frame pluralistic alignment as a multi-objective RL problem \citep{hayes2022practical,rame2023soups}, where we train a multi-objective LM (see \cref{fig:figure1} left) that can be steered to generate desirable outputs over the continuum of possible reward weightings. Specifically, the multi-objective LM takes as input the prompt $x$ \emph{and} reward weighting $w$ and responds with outputs $y\sim\pi_{\op{MO}}(\cdot\mid x,w)$ that maximize the specified reward combination $\sum_jw_jr_j$ in expectation (up to KL-regularization). In contrast, the classic single-objective LM only takes input the prompt $x$ and outputs $y\sim\pi_{\op{SO}}(\cdot\mid x)$ that maximize a fixed scalar reward, such as $\sum_j w^{\op{fix}}_jr_j$ for some fixed weightings vector $w^{\op{fix}}$.
An important application of multi-objective language models is \emph{decision support}: a system which can provide multiple diverse generations that support or cover a wide space of interests and lets the user pick their favorite \citep{roijers2016multi}, which can in turn be used to personalize the model to the user.

In the context of LMs, multi-objective finetuning (MOFT) has been explored via prompt-based approaches \citep{jang2023personalized,guo2024cpo,wang2024dpa} and parameter-based approaches \citep{rame2023soups,jang2023personalized}.
Prompt-based approaches finetune a LM that is steered by simply including the reward weightings into the prompt. However, prompt-based MOFT is \emph{sub-optimal in steerability} as we show in our experiments and can be sensitive to how weightings are encoded in the prompt.
An alternative parameter-based approach, Rewarded Soups (RS) \citep{rame2023soups}, (a) \emph{independently} trains one LM per reward function, and (b) interpolates parameters from separate LMs with reward weightings to perform conditional generation at inference-time.
Surprisingly, this \emph{zero-shot} approach can effectively trade-off multiple rewards by relying on the linear mode connectivity \citep{frankle2020linear}. However, we find that zero-shot MOFT is \emph{sub-optimal in performance} on intermediate weightings that it hasn't seen during training. %

This paper presents Conditional Language Policy (\clp), a general framework for that employs parameter-space conditioning with multi-task training \citep{dosovitskiy2020you} and achieves state-of-the-art results in our extensive experiments on summarization. Using parameter-conditioning from RS, \clp is \emph{consistently more steerable} than purely prompt-based approaches.
Moreover, by finetuning on a diverse set of reward weightings, \clp produces \emph{higher quality responses} than zero-shot approaches like RS while having comparable or better steerability.
We conduct a systematic set of experiments, observing that \clp both Pareto-dominates RS and is more steerable than prompt-based MOFT (see \cref{fig:figure1} right). \clp robustly maintains these benefits across experimental conditions and across various reward functions and model sizes.
We also conduct an automated evaluation with Gemini 1.0 Ultra \citep{team2023gemini} that further supports that \clp is more steerable and generates higher quality responses than existing baselines. Finally, we provide novel theory proving that zero-shot methods can display near-Pareto-optimal behavior when optimal policies for individual rewards are aligned, but they provably fail otherwise. In those failure cases, multi-task training (as used by \clp) is necessary to learn to Pareto-optimal policy.

In summary, our contributions are:
\begin{enumerate}[itemsep=0.3ex,topsep=0ex,partopsep=0ex,parsep=0ex]
  \item We propose \clp, a general framework for MOFT that learns multi-objective LMs through multi-task learning and parameter efficient model averaging (\cref{sec:moft-with-clp}).
  \item We extensively evaluate \clp on summarization and show it robustly improves existing approaches in \emph{both} output quality and steerability, across many experimental conditions and automated evaluator evaluations (\cref{sec:experiments}).
  \item We theoretically prove that logit mixing, a special case of \clp, is near-optimal under a coverage condition. We also provide an example instance where zero-shot (\ie, without multi-task training) methods fail while \clp succeeds (\cref{sec:theory}).
\end{enumerate}

\section{Problem Setup}\label{sec:problem-setup}
Let $\piref(y\mid x)$ be a base policy with parameters $\thetaref$, where $x$ is the input prompt and $y$ is the output generation.
In \textbf{single-objective finetuning} (SOFT), there is a fixed reward function $R(x,y)$ and the goal is to maximize the expected reward without drifting too far from $\piref$. Formally, SOFT learns a policy $\pi_\theta(\cdot)$ to maximize the following value:
\begin{equation}
  V_{\alpha,R}(\pi):=
  \EE_{x\sim\Dcal,y\sim\pi(x)}[(1-\alpha)R(x,y)
  -\alpha\KL(\pi(\cdot\mid x)\Mid\piref(\cdot\mid x)) ], \label{eq:single-objective-ft}
\end{equation}
where $\alpha\in(0,1)$ is a KL weighting that controls the deviation from $\piref$.\footnote{Recall for distributions $P\ll Q$, the KL-divergence is defined as $\KL(P\Mid Q):=\EE_{P}[\log\frac{\diff P}{\diff Q}]$. If $P\not\ll Q$, KL is $\infty$.}
SOFT can be solved by standard RL algorithms such as REINFORCE \citep{williams1992simple} or PPO \citep{schulman2017proximal}.
Notice that the KL-term can be thought of as a special reward function to mitigate reward-hacking \citep{stiennon2020learning}.
The main issue with SOFT is that both the reward function $R$ and the KL weightings are fixed and unchangeable after training; thus, it is not possible to offer near-optimal behavior on other reward and KL weightings without retraining SOFT with multiple reward or $\alpha$-values.

\textbf{Multi-objective finetuning} (MOFT) fixes the above issues by learning a {\em conditional} policy that generates $y$ based not only on the prompt $x$, but also the $\alpha$ and $R$ specified at test-time \emph{without retraining}.
We formulate MOFT using multi-objective RL \citep{roijers2016multi} and consider a vectorial reward $\rvec(s,a)\in\RR^m$ whose linear scalarizations $\{w^\top\rvec(\cdot):w\in\Delta_m\}$ capture all the desired trade-offs at test-time, where $\Delta_m$ is the $(m-1)$-simplex, \ie, $\rvec$ is a basis for all desired rewards at test-time.
The goal is to learn parameters $\phi$ such that, for all \emph{weightings} $\alpha\in[\alphamin,1]$ and $w\in\Delta_m$,
the \emph{conditioned} policy $\pi_\phi(\cdot;\alpha,w)$ maximizes $V_{\alpha,w^\top\rvec}$, the objective with KL-regularizer $\alpha$ and reward function $w^\top\rvec(\cdot)$.
We frame MOFT as multi-task training over the weighting distribution $\Qcal$, and aim to maximize:
\begin{equation}
  V_{\text{moft}}(\phi) = \EE_{(\alpha,w)\sim\Qcal}[V_{\alpha,w^\top\rvec}(\pi_\phi(\cdot;\alpha,w))]. \label{eq:moft-mtl-obj}
\end{equation}
In theory, the optimal solution to multi-objective RL is a large (potentially exponential in $m$) set of policies called the convex coverage set \citep{roijers2016multi}.
However, the rich representational power of a single LLM may already be able to approximate such a policy cover and thus we aim to solve MOFT with a memory-efficient parameterization $\phi$. %

\begin{algorithm}[!t]
\caption{\clp: Conditional Language Policy}\label{alg:CLP-framework}
\textbf{Input:} Weightings sampler $\Qcal$, number of training steps $T$, learning rates $\{\eta_t\}_{t=1}^T$.
\begin{algorithmic}[1]
\State Init \clp: $\phi_0 = (\thetaref[\Scal^C],\{\thetaref[\Scal]\}_{i\in[m]})$.
\For{$t$ in $0 \dots T-1$}
    \State Sample prompt $x_t\sim\Dcal$ and KL \& reward weightings $(\alpha_t,w_t)\sim\Qcal$. \label{line:sample-prompt-weights}
    \State Get conditioned policy $\pi_t(\cdot;\alpha_t,w_t)$ from running \cref{alg:conditioning-mechanism} with $\alpha_t,w_t,\phi_t$.
    \State Conditioned generation $y_t\sim \pi_t(x_t;\alpha_t,w_t)$ (same computation as single LM). \label{line:compute-conditioned-generations}
    \State Objective: $r_t\gets (1-\alpha_t)\cdot w_t^\top \rvec(x_t,y_t) - \alpha_t\cdot\KL(\pi_t(\cdot\mid x_t;\alpha_t,w_t)\Mid\piref(\cdot\mid x_t))$.\label{line:compute-objective}
    \State Update \clp parameters: $\phi_{t+1} \gets \phi_t + \eta_t\cdot g_t$ where $g_t= r_t\cdot\nabla_{\phi_t}\log\pi_t(y_t\mid x_t;\alpha_t,w_t)$. \label{line:update-phi-parameters}
\EndFor\\
\textbf{Output:} \clp parameters $\phi_T$.
\end{algorithmic}\label{alg:CLP-training}
\end{algorithm}

\textbf{MOFT Desiderata -- Pareto-dominance \& steerability.}
A multi-objective LM $\pi$ \emph{Pareto-dominates} another $\pi'$ if $V_{\alpha,w^\top\rvec}(\pi(\cdot;\alpha,w))\geq V_{\alpha,w^\top\rvec}(\pi'(\cdot;\alpha,w))$ for all values of $\alpha,w$ that one cares about.
New MOFT algorithms should ideally Pareto-dominate existing baselines to ensure that generation quality is improved along all axes.
\emph{Steerability} is another important goal for MOFT algorithms. In \cref{fig:figure1}, \clpall and \clpp both satisfy the first goal of Pareto-dominance, but \clpall is desirable since its Pareto-curve has much better spread, \ie, it is more steerable.

\textbf{Notation.}
$\theta$ refers to LM parameters, while $\phi$ refers to $\clp$ parameters (different structure from $\theta$), which can be conditioned on $(\alpha,w)$ to produce a conditioned LM parameter $\theta^{\alpha,w}$ (same structure as $\theta$).
$\theta[\Scal]$ or $\theta_\Scal$ refer to the subset of parameters indexed by $\Scal$. We use $\oplus$ to combine disjoint parameter subsets, \ie, $\theta=\theta[\Scal^C]\oplus\theta[\Scal]$. We assume $\theta$ and $\thetaref$ have the same structure.

\begin{algorithm}[!t]
\caption{Conditioning Mechanism}
\textbf{Fixed:} KL-mixing function $\mixmap(\cdot)$, index of conditioning params $\Scal$, $\CondPrompt = \textsc{False}$. \\
\textbf{Input:} KL \& reward weightings $(\alpha,w)$, \clp parameters $\phi=(\theta_{\Scal^C},\{\theta_\Scal^{(i)}\}_{i\in[m]})$.
\begin{algorithmic}[1]
\State Set $\beta=\mixmap(\alpha)$ and compute:
\begin{equation}
  \theta_\Scal^{\alpha,w}
  \gets \textstyle(1-\beta)\sum_{i=1}^m w[i]\cdot\theta_{\Scal}^{(i)}
  +\beta\cdot\thetaref[\Scal]. \label{eq:parameter-conditioning}
\end{equation}
\State Combine params $\theta^{\alpha,w}=\theta_\Scal^{\alpha,w}\oplus \theta_{\Scal^C}$.
\State \textbf{Return:} Policy that concatenates weights to input $\{x\mapsto \pi_{\theta^{\alpha,w}}(\texttt{CONCAT}([w, x]))\}$ if \CondPrompt. Else, return policy $\pi_{\theta^{\alpha,w}}$ \label{line:clp-output}
\end{algorithmic}\label{alg:conditioning-mechanism}
\end{algorithm}

\section{Conditional Language Policy (CLP)}\label{sec:moft-with-clp}
This section presents the Conditional Language Policy (\clp) framework for MOFT, which enables a family of algorithms with varying trade-offs between quality (in terms of Pareto-dominance and steerability) and cost (in terms of parameter-count). In brief, \clp learns a set of parameters $\phi$ that can be processed into a {\em conditioned} LM for any given weighting across rewards and KL, via a parameter-averaging mechanism described in \cref{sec:def-conditioning-mechanism}. The learning algorithm samples a diverse set of weightings to push out its Pareto-front over all weightings simultaneously. Notably, this is multi-task learning across the continuum of weightings, which directly maximizes the MOFT objective defined in \cref{eq:moft-mtl-obj}, unlike the existing zero-shot approaches  \citep{rame2023soups,jang2023personalized}.

We describe the \clp algorithm in \cref{alg:CLP-training} (illustrated in \cref{fig:clp}), where each training round $t=1,2,\dots,T$ consists of three steps.
First, we sample a prompt $x_t$ and reward \& KL weightings  $w_t,\alpha_t\sim\Qcal$ for this round (\cref{line:sample-prompt-weights}).
Second, we condition \clp on weightings $(\alpha_t,w_t)$ and sample generations $y_t\sim \pi_t(x_t;\alpha_t,w_t)$ (\cref{line:compute-conditioned-generations}).
Third, we compute the conditioned objective (\cref{line:compute-objective}) and update the \clp parameters with policy gradient (\cref{line:update-phi-parameters}).
The policy optimization step uses gradient ascent to maximize the objective in \cref{eq:moft-mtl-obj} and can be implemented by any standard RLHF algorithms such as REINFORCE \citep{williams1992simple,ahmadian2024back}, PPO \citep{schulman2017proximal}, and also DPO \citep{rafailov2023direct} when given pluralistic preference data.

\subsection{Conditioning Mechanism}\label{sec:def-conditioning-mechanism}
We now describe the parameter-based mechanism for computing conditional policies $\pi(\cdot;\alpha,w)$ in \cref{alg:conditioning-mechanism}.
Let $\Scal$ denote an index set on the LM parameters that we wish to use for parameter-conditioning \citep{rame2023soups}, \eg, attention weights, and its choice can trade-off the steerability and memory cost of \clp.
We maintain (1) $m$ sets of conditioned parameters $\{\theta_\Scal^{(i)}\}_{i\in[m]}$ indexed by $\Scal$, and (2) one set of unconditioned parameters $\theta_{\Scal^C}$.
To condition the $\Scal$-part on $(\alpha,w)$, we linearly combine the $m$ conditioning parameters with weightings $w$, and we then combine the result with $\thetaref[\Scal]$ weighted by $\mixmap(\alpha)$ (\cref{eq:parameter-conditioning}).
Then, we concatenate the conditioned part $\theta^{\alpha,w}_\Scal$ with the unconditioned part to obtain the full LM parameters $\theta^{\alpha,w}=\theta_\Scal^{\alpha,w}\oplus\theta_{\Scal^C}$. The parameter count of \clp is thus $\Ocal(m|\Scal|+|\Scal^C|)$.
We note that inference with \clp only requires one forward pass through the LM and the above parameter-averaging cost is amortized since it is done only once at the beginning. See \cref{app:gradient-analysis} for details on gradient propagation through condtioning.

\noindent\textbf{Prompt-based conditioning}: One can also augment the prompt with the reward weightings (\cref{line:clp-output} in \cref{alg:conditioning-mechanism}); see also \cref{app:prompt-based} for more details on prompt-design.
Prompting based MOFT has been explored in recent manuscripts \citep{jang2023personalized,guo2024cpo,wang2024dpa} and has the advantage of not requiring additional parameters, but consumes part of the context and is sensitive to how these are encoded in the context. In \cref{sec:clp-with-prompt-conditioning}, we consider augmenting parameter-space conditioning with prompting.

\subsection{Three Instantiations of CLP}\label{sec:parameter-conditioning-instantiations}
The choice of $\Scal$ influences both the steerability and memory usage of \clp.
On one extreme, the most steerable and high parameter count choice is to condition on all LM parameters and we call this \clpall, \ie, $\Scal_{\text{full}}=\{\text{indices of all LM parameters}\}$. This instance is inspired by model soups \citep{wortsman2022model}.
On the other extreme, \clpl only conditions on the final linear layer (a.k.a. logit layer), \ie, $\Scal_{\text{logit}}=\{\text{indices of last linear layer of LM}\}$. This instance is theoretically grounded \citep{liu2024decoding} but we found it to have inferior steerability.
A nice compromise is \clpa which conditions on the attention parameters of the LM, \ie, $\Scal_{\text{attn}}=\{\text{indices of attention layers of LM}\}$. \clpa is much more parameter-efficient than \clpall and we find that it is nearly as steerable as \clpall in our experiments.

In our experiments, we observe that the more expressive parameterizations of $\Scal$ (\eg, $\Scal_{\text{full}}$ and $\Scal_{\text{attn}}$) robustly lead to Pareto-dominating and highly steerable behaviors than existing baselines such as Rewarded Soups. We remark that expressivity is determined not just by the number of parameters in $\Scal$ but also where those parameters are in the LM ({\em e.g.} earlier vs. later layers).
Finally, we highlight that \clp is agnostic to the LM architecture and $\Scal$ can be set appropriately for any model type.

\section{Experiments}\label{sec:experiments}
We consider the following questions:
\begin{itemize}[leftmargin=*]
    \setlength\itemsep{0em}
    \item {\bf Benchmarking:} How do different methods perform in terms of \emph{performance} (ability to push out the Pareto Front) and \emph{steerability} (ability to generate content that trades-off different objectives)?
    \item {\bf Ablations:} How does the behavior of different approaches vary as a function of (a) number of finetuning steps, (b) model size? Furthermore, is parameter space conditioning composable with prompting based methods?
    \item {\bf Automated Evaluations:} Going beyond Pareto fronts, we present automated evaluations that compare generations from \clp against baselines by having Gemini~\citep{team2023gemini} rate the summaries in terms of quality and steerability.
\end{itemize}
\noindent\textbf{Data and models.} Most of our experiments and ablations are performed on summarization with the widely-used XSum dataset~\citep{narayan-etal-2018-dont}.
We initialize the reference policy $\piref$ and reward models from the instruction finetuned (FLAN) checkpoints for T5 \citep{chung2024scaling}.
We use the large size (770M parameters) for reward models and we mostly use the base size (220M parameters) for policies, except in our model size ablation where we also use large size for policies. In addition to these main results, we also ran experiments on the TLDR dataset~\citep{volske2017tl} with larger policy (T5-XL size) and reward models (T5-XXL size) to show that our general observations are scalable and robust; these experiments are presented in \cref{ssec:tldr-dataset-results}.

\noindent\textbf{Reward functions.} We consider three reward functions:
(1) ROUGE, a formulaic reward that measures similarity of generation to the ground truth summary \citep{lin2004rouge};
(2) natural language inference (NLI), a learnt reward model for textual entailment and factuality \citep{nie-etal-2020-adversarial, roit2023factually};
and (3) a reward model for summary quality learnt from the ``too long; didn't read'' (TLDR) dataset \citep{stiennon2020learning}.
ROUGE and the quality model (referred to as TLDR) tend to favor verbose and descriptive summaries while NLI favors concise summaries; this gives our setup a distinct tension between various reward pairs.

\noindent\textbf{Methods.}
We benchmark the three instances of \clp in \cref{sec:parameter-conditioning-instantiations}: \clpall, \clpa, \clpl. As a gentle reminder, \clpall maintains one full LM per reward, akin to standard model soups; \clpa maintains replicas of all attention layers thus being more parameter efficient; and \clpl maintains replicas only the logit layer which is theoretically grounded but less expressive. For policy optimization, we use REINFORCE with control variate, which is a lighter implementation than PPO \citep{schulman2017proximal} and has been successfully used for summarization \citep{roit2023factually}.

\noindent\textbf{Baselines.} We consider (a) Rewarded Soups (RS)~\citep{rame2023soups} which independently trains one policy per reward and linearly interpolates the parameters with weightings at inference time -- this is a `zero-shot' version of \clpall;
(b) a \clpp baseline~\cite{jang2023personalized, guo2024cpo}, which encodes the reward weightings into the prompt and is trained with our multi-task objective -- for details on the prompt and how it was selected, see \cref{app:prompt-based}.
For the ``single-reward, multi-KL'' setting (\cref{sec:single-reward-varying-kl}), instead of RS, we consider (c) the recent Decoding-time Realignment (DeRA)~\citep{liu2024decoding}, which maintains two LMs (an LM optimized for $\alpha_{\min}$ and reference LM $\piref$) and linearly interpolates their logits.

All methods and baselines are run for \emph{same number of training iterations}. See \cref{app:more-training-details} for related details and hyper-parameters.

\noindent\subsection{Core Benchmarking Results}
\subsubsection{Single Reward, Multi KL Regularizer}\label{sec:single-reward-varying-kl}
\begin{figure}[!h]
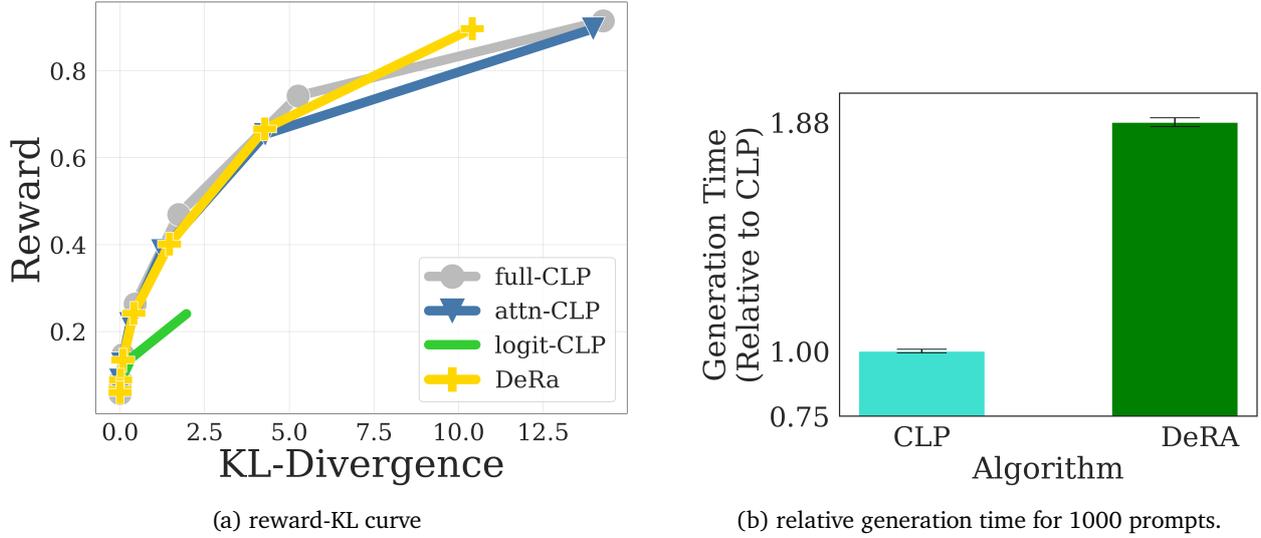

  \begin{subfigure}{0.5\linewidth}
  \includegraphics[width=\linewidth]{without_so_plots/so_softmdp_reward_v_kl_alpha_0_01.pdf}
  \caption{reward-KL curve} \label{fig:single-objective-kl-pareto-front}
  \end{subfigure}\hspace*{\fill}
  \begin{subfigure}{0.45\linewidth}
  \includegraphics[width=\linewidth]{misc_plots/generation_time.pdf}
  \caption{relative generation time for $1000$ prompts.}\label{fig:timing-comparison-clp-DeRA}
  \end{subfigure}
  \caption{Pareto curves for single-reward, multi-$\alpha$. Observe \clp variants (\clpall and \clpa) are competitive with DeRA, a baseline that is nearly $2\times$ expensive to run at inference time.}
\end{figure}

In the first setting, we fix a single reward function and vary the KL regularizer $\alpha$ to test the KL-reward trade-off. We use the reward $R=0.9\RNLI + 0.1\RRouge$, where $\RRouge$ is mixed in to mitigate reward hacking the NLI model.
In \cref{fig:single-objective-kl-pareto-front}, we see that all methods except \clpl are able to evenly trade-off reward and KL, enabling a smooth transition from $\piref$ (when $\alpha=1$) to a maximally finetuned model (when $\alpha=0.01$).
This suggests that parameter-mixing trained with the multi-task objective is competitive with the baseline DeRa, which is state-of-the-art for reward-KL trade-off.
Moreover, \cref{fig:timing-comparison-clp-DeRA} shows that \clp is also $\sim 2\times$ more computationally efficient than DeRa at inference-time, because DeRa performs two LM calls (both $\piref$ and $\pi_{\alphamin}$) per token.
Hence, inference speed is a major benefit of parameter-mixing over logit-ensembling.

\subsubsection{Two Rewards, Fixed KL Regularizer}\label{sec:two-rewards-fixed-kl-regularizer}
\begin{figure}[h!]
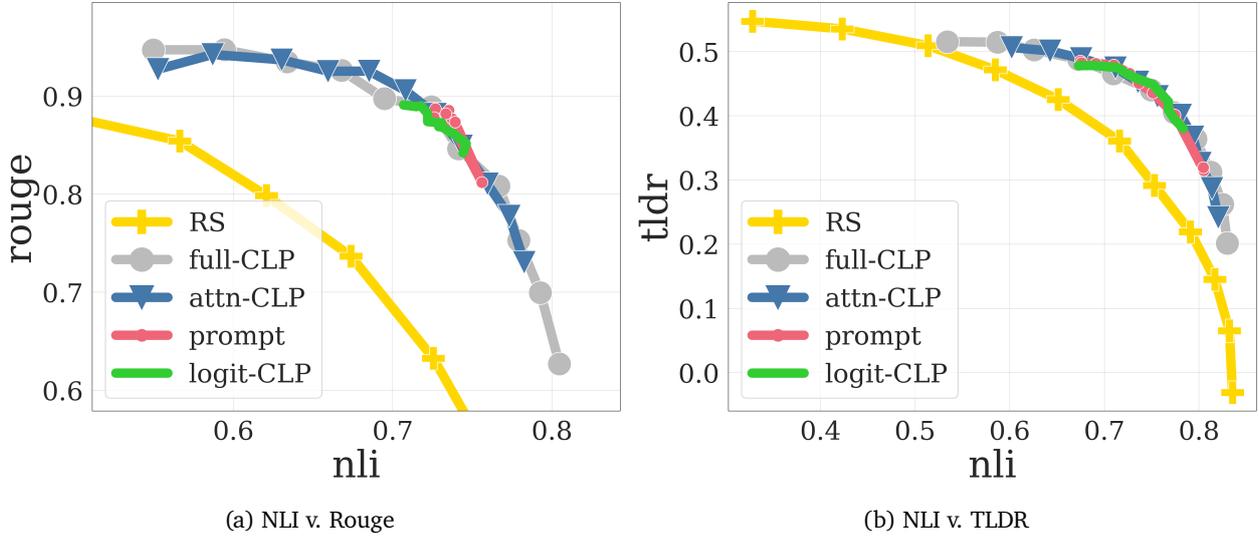

  \begin{subfigure}{0.5\linewidth}
  \includegraphics[width=\linewidth]{without_so_plots/abridged_base_nli_rouge_alpha_0_01.pdf}
  \caption{NLI v. Rouge} \label{fig:two-reward-nli-rouge}
  \end{subfigure}\hspace*{\fill}
  \begin{subfigure}{0.5\linewidth}
  \includegraphics[width=\linewidth]{without_so_plots/base_nli_tldr_alpha_0_01.pdf}
  \caption{NLI v. TLDR} \label{fig:two-reward-nli-tldr}
  \end{subfigure}
  \caption{Pareto-curves for two-reward \& $\alpha=0.01$. Observe \clp variants (\clpall and \clpa) offer improved spread (compared to \clpp) while Pareto-dominating the Rewarded Soups (RS) baseline.} \label{fig:two-reward-fixed-alpha}
\end{figure}

Here, we consider a pair of rewards with a fixed KL regularizer, to test the trade-off between both rewards.
\cref{fig:two-reward-fixed-alpha} shows the Pareto curves for (a) NLI v. Rouge and (b) NLI v. TLDR, where the x,y-axes are the KL-regularized rewards, \ie, $x=V_{\alpha,R_1}(\pi(\cdot;\alpha,w)),y=V_{\alpha,R_2}(\pi(\cdot;\alpha,w))$, which recall is the true objective being maximized by finetuning (\cref{eq:single-objective-ft}).
We see that \clp and \clpp largely Pareto-dominate the baseline RS, which shows the benefits of multi-task training compared to RS's zero-shot approach.
\clpall and \clpa both exhibit Pareto-fronts that are more steerable (evenly spaced and spread out) than those of \clpl and \clpp, which largely exhibit a mode-collapsed behavior.
Importantly, while efficiently replicating only the attention weights, \clpa can Pareto-dominate the baseline RS while maintaining steerable Pareto-curves.
Thus, \clpa offers the best trade-off between steerability and parameter count.
See \cref{app:base-two-reward-fixed-alpha} for more results.

\subsubsection{Three Rewards, Fixed KL Regularizer}
\begin{figure}[!h]
  \centering
  \includegraphics[width=\linewidth]{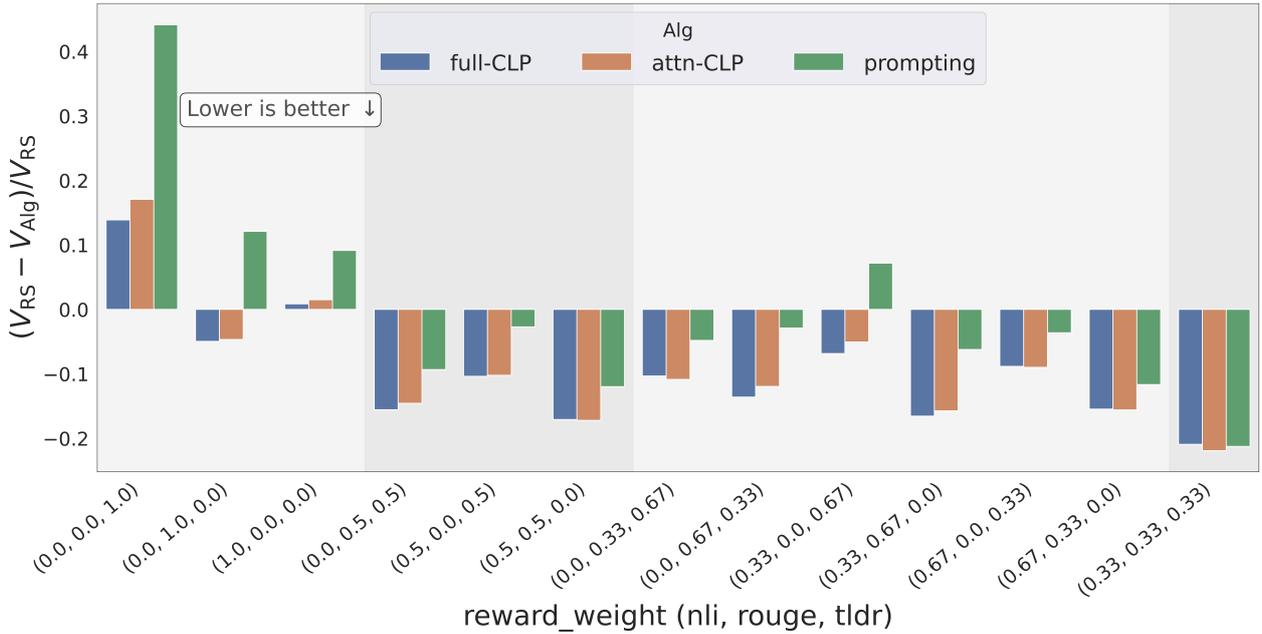}
  \caption{Barplot of $1.0-V_{w^\top\rvec}(\pi_{\text{Alg}}(\cdot;w))/\wt V_{w^\top\rvec}^{\text{RS}}$ for three-reward experiments, where $\wt V_{w^\top\rvec}^{\text{RS}}$ is the KL-regularized reward of RS for weighting $w$. Lower is better and $0$ is on-par with RS.}
  \label{fig:three-rewards-full-attn-clp}
\end{figure}

We now consider the three-reward setting to test the trade-off between all reward functions.
\cref{fig:three-rewards-full-attn-clp} plots the KL-regularized value $V_{\alpha,w^\top\rvec}(\pi(\cdot;\alpha,w))$, normalized w.r.t. RS, for $13$ different reward weightings $w$ shown below the $x$-axis. At the extreme weights $w=e_i,i\in\{1,2,3\}$, RS is equivalent to single-objective finetuning and naturally outperforms \clp and \clpp as expected.
At the intermediate weights $w\neq e_i$, \clpall consistently outperforms RS.
We find that \clpa is competitive to \clpall despite incurring only $20\%$ of the memory overhead.
Thus, akin to the two-reward setting, \clpa achieves the best balance in terms of steerability and parameter count. %
For additional three reward results, see \cref{app:base-three-reward-fixed-alpha}.

\noindent\textbf{Summary of core benchmarking results.} In terms of performance, we find that multi-task training enables \clp to improve over the zero-shot RS baseline. Importantly, we find that \clpall and \clpa robustly maintain a steerable Pareto-front that is more spread out than \clpl and \clpp baseline. In sum, \clpa presents a favorable trade-off in terms of Pareto-front and steerability, while using fewer parameters than existing baselines.

\subsection{Ablation Studies}
\subsubsection{Effect of Training Iterations}\label{sssec:training_iterations}
\begin{figure}[h!]
\includegraphics[width=\linewidth]{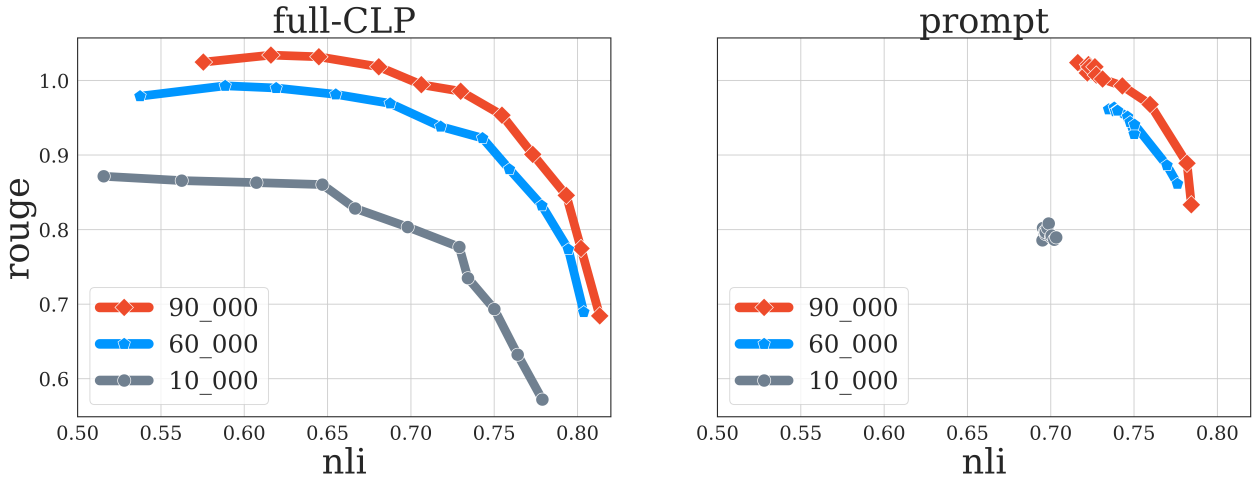}
\caption{Pareto-curves at $10k,60k,90k$ training steps. Observe that \clpp shows slightly improved steerability with a $3\times$ larger training budget but still isn't as steerable as \clpall which exhibits a strong steerability even at $10k$ iterations.}
\label{fig:long-run-ablation}
\end{figure}

\cref{fig:long-run-ablation} shows the progression of Pareto-curves over $90k$ training steps for \clp instances.
\clpall and \clpa are steerable after just $10k$ steps. In contrast, \clpp becomes steerable at $60k$ steps; \clpl has a similar trend of being not as steerable at earlier training iterations. %
Furthermore, the optimization dynamics of \clpall and \clpa are much smoother than \clpp and \clpl. Complete results are presented in \cref{app:ablation-training-iterations}.

\subsubsection{CLP With Prompt Conditioning}\label{sec:clp-with-prompt-conditioning}
\begin{figure}[h!]
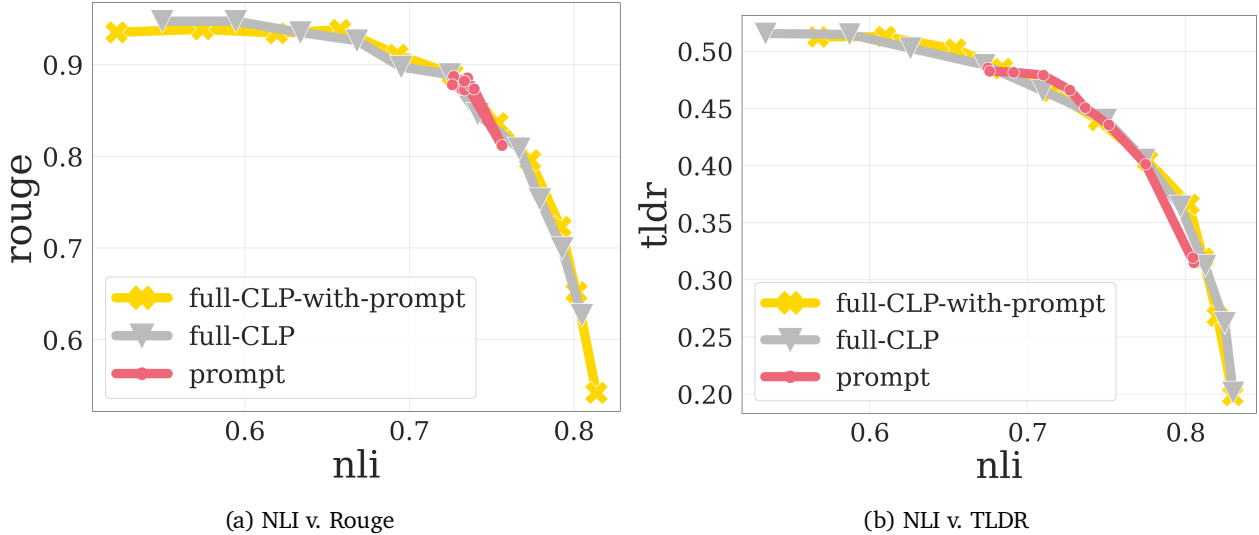

  \begin{subfigure}{0.5\linewidth}
  \includegraphics[width=\linewidth]{prompt_ablation_and_longrun/full_CLP_base_nli_rouge_alpha_0_01.pdf}
  \caption{NLI v. Rouge}
  \end{subfigure}\hspace*{\fill}
  \begin{subfigure}{0.5\linewidth}
  \includegraphics[width=\linewidth]{prompt_ablation_and_longrun/full_CLP_base_nli_tldr_alpha_0_01.pdf}
  \caption{NLI v. TLDR}
  \end{subfigure}
  \caption{Combining Prompt-based conditioning with Parameter-space condtioning for \clpall. Observe that \clpp slots in with \clpall to produce steerable and Pareto dominating behaviors.} \label{fig:prompt-ablation}
\end{figure}

\cref{fig:prompt-ablation} shows the Pareto-fronts of \clpall, the \clpp baseline, and \clpall with prompting ($\CondPrompt=\textsc{True}$).
For NLI v. Rouge, \clpall with prompting showed a slight improvement in steerability, whereas there was little difference for NLI v. TLDR.
Hence, prompt conditioning does not hurt performance but can be slightly beneficial.
See \cref{app:combining-prompt-parameter-conditioning} for more results.

\subsubsection{Model Size}\label{sec:ablation-model-size}
\begin{figure}[h!]
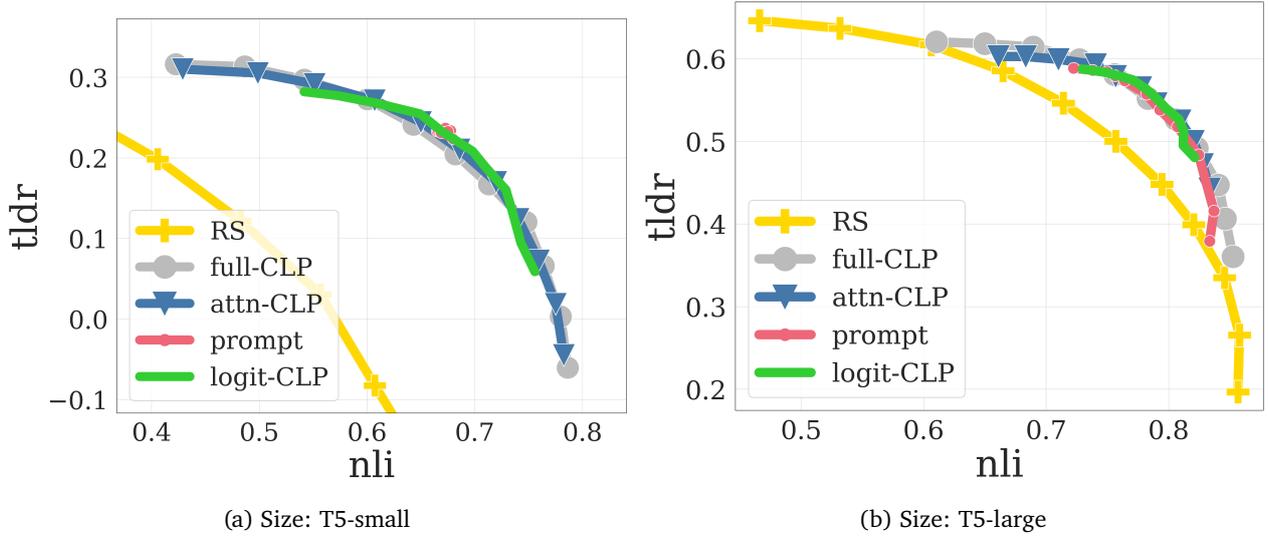

  \begin{subfigure}{0.5\linewidth}
  \includegraphics[width=\linewidth]{without_so_plots/abridged_small_nli_tldr_alpha_0_01.pdf}
  \caption{Size: T5-small} \label{fig:nli-tldr-t5-small}
  \end{subfigure}\hspace*{\fill}
  \begin{subfigure}{0.5\linewidth}
  \includegraphics[width=\linewidth]{without_so_plots/large_nli_tldr_alpha_0_01.pdf}
  \caption{Size: T5-large} \label{fig:nli-tldr-t5-large}
  \end{subfigure}
  \caption{Ablation on model size for NLI v. TLDR. Observe that across model sizes, \clpall and \clpa learn steerable (compared to \clpp) and Pareto-dominating behaviors (compared to Rewarded Soups).} \label{fig:model-size-ablation}
\end{figure}

We rerun our experiments with T5-small (60M) and T5-large (770M) to check how our findings change with different model sizes.
\cref{fig:model-size-ablation} shows the NLI v. TLDR Pareto-fronts. We see that \clpall and \clpa still robustly Pareto-dominate the RS baseline and maintain steerable fronts.
Interestingly, \clpp collapses to a point for T5-small but has much better spread for T5-large, suggesting that \clpp is sensitive to model size and a larger model can improve the steerability of \clpp. We provide results for other reward pairs in \cref{app:ablation-model-size}.

\noindent\textbf{Summary of ablations.}
While combining \clpp with \clp did not significantly improve steerability, \clpp may exhibit more steerability with larger models or training time.
Across different settings, \clpall and \clpa consistently maintain their superior performance and steerability suggesting they are robust conditioning architectures.

\subsection{Automated Evaluation}
\begin{figure}[h!]
  \begin{subfigure}{0.5\linewidth}
  \includegraphics[width=\linewidth]{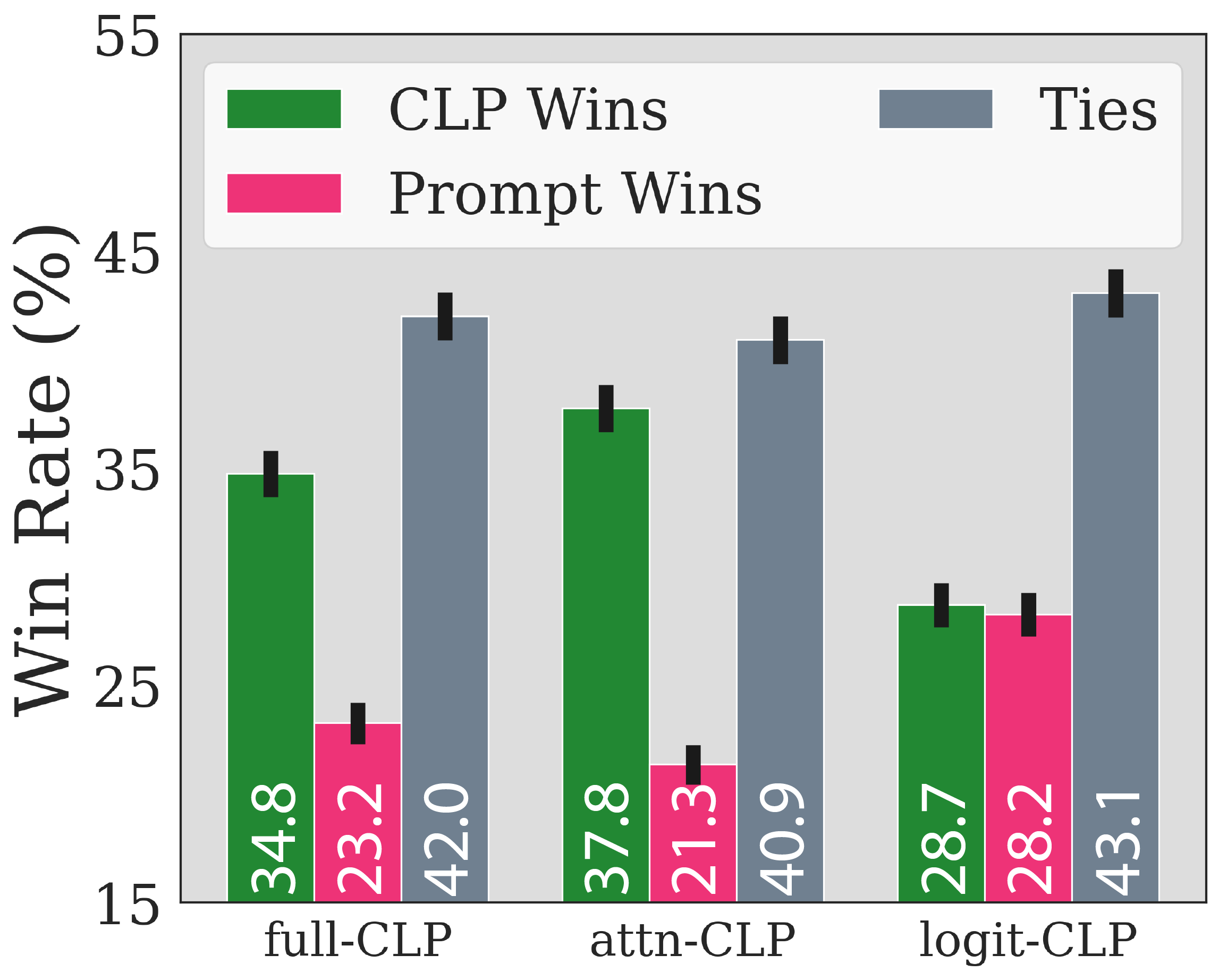}
  \caption{\clp vs. \clpp} \label{fig:clp-vs-prompt-automated_evaluator}
  \end{subfigure}\hspace*{\fill}
  \begin{subfigure}{0.5\linewidth}
  \includegraphics[width=\linewidth]{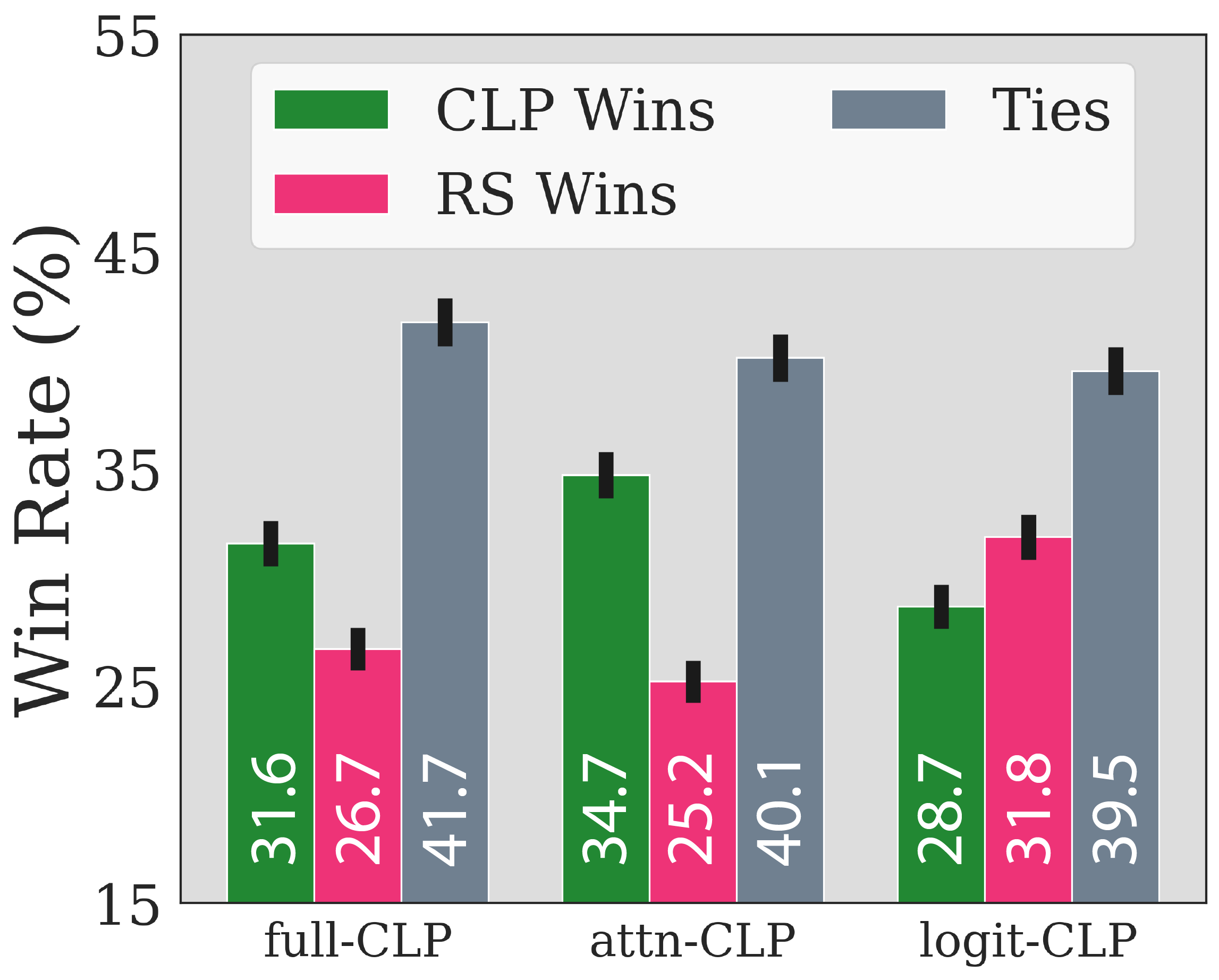}
  \caption{\clp vs. Rewarded Soups} \label{fig:clp-vs-rs-automated_evaluator}
  \end{subfigure}
  \caption{Automated Evaluation Win Rate comparison of \clp variants against prompting and Rewarded Soups baselines. \clp variants (\clpall and \clpa) improve win rates by 5 to 10 (and 10 to 15) \% compared to Rewarded-Soups (and \clpp) baselines.} \label{fig:automated_evaluator-win-rates}
\end{figure}

In order to understand if \clp's improved Pareto-fronts translate to improvements in generations compared to baselines (\clpp and Rewarded Soups), we conduct an automated evaluation of generation quality and steerability. For this evaluation, we consider the NLI v. TLDR setup with T5-large models (from \cref{sec:ablation-model-size}) and use $2000$ articles from the XSum validation set.
We utilize Gemini 1.0 Ultra~\citep{team2023gemini} as an automated evaluator to compare summaries from \clp instances to each baseline in terms of their conciseness and summary quality.
Specifically, for each article, we sample conditioned summaries from both \clp and a baseline on weightings $w=(0.8,0.2)$ for high NLI (resp. weightings $w=(0.2,0.8)$ for high TLDR) and we ask the automated evaluator to compare which summary is more concise (resp. has higher quality).
We permute the comparison order to account for position bias \citep{wang2023large}, marking a comparison as \emph{consistent} if both permutations agree -- please see \cref{app:details-autorater} for details.
Then, for each article, we consider an algorithm to be the winner (\ie, more steerable) if either the auto evaluator prefers its summary in both comparisons, or the auto evaluator prefers its summary in one case and was inconsistent in the other. If neither algorithm is the winner, we consider it a tie.
With this setup in place, \cref{fig:automated_evaluator-win-rates} shows the win-rate across different \clp variants against the two baselines.
We observe that \clpall (and \clpa) offers $11.6\%$ (and $16.5\%$) improvement in raw win rates compared to the multi-task trained \clpp baseline, and $4.9\%$ (and $9.5\%$) improvements over RS. \clpl tends to fair comparably to \clpp while being inferior to RS (dropping win rate by $3.1\%$). Notably, \clpa and \clpall achieve the best win-rate relative to both baselines in this automatic evaluation, with \clpa having an additional desirable property of being more parameter-efficient. In sum, our automatic evaluation is consistent with prior Pareto-front results and validates that \clp produces higher quality multi-objective LMs with superior steerability both quantitatively and qualitatively.

\subsection{Scaling up \clp with larger policy and reward models}\label{ssec:tldr-dataset-results}
\begin{figure}[t!]
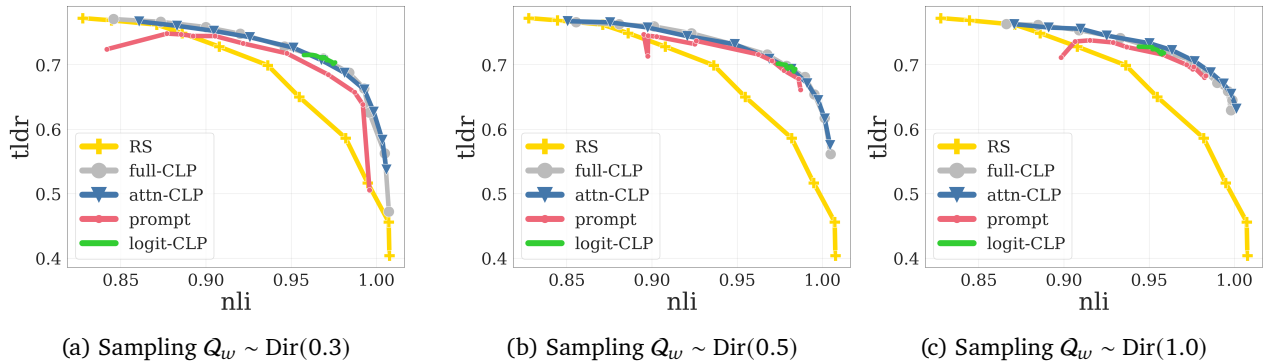

  \begin{subfigure}{0.32\linewidth}
  \includegraphics[width=\linewidth]{misc_plots/tldr_dataset_dir_03xl_nli_tldr_alpha_0_01_6k.pdf}
  \caption{$\Qcal_w\sim \text{Dir}(0.3)$} \label{fig:tldr-results-dir03}
  \end{subfigure}\hspace*{\fill}
  \begin{subfigure}{0.32\linewidth}
  \includegraphics[width=\linewidth]{misc_plots/tldr_dataset_dir_05xl_nli_tldr_alpha_0_01_6k.pdf}
  \caption{$\Qcal_w\sim \text{Dir}(0.5)$} \label{fig:tldr-results-dir05}
  \end{subfigure}
  \begin{subfigure}{0.32\linewidth}
  \includegraphics[width=\linewidth]{misc_plots/tldr_dataset_dir_10xl_nli_tldr_alpha_0_01_6k.pdf}
  \caption{$\Qcal_w\sim \text{Dir}(1.0)$} \label{fig:tldr-results-dir10}
  \end{subfigure}
\caption{Results on the TLDR dataset with T5-XL policy and T5-XXL reward models. These experiments vary the sampling distribution $\Qcal_w$ of reward weightings (Dirichlet$(0.3)$ is more narrow while Dirichlet$(1.0)$ is uniform). We observe that the steerability of different methods improve as $\Qcal_w$ becomes narrower, though this may come at the cost of a slightly inferior Pareto front.} \label{fig:tldr-results}
\end{figure}

In the following experiment, we perform multi-reward finetuning on the TLDR dataset \citep{volske2017tl,stiennon2020learning} and increase the model size of both the policy and reward models. Namely, we train two reward models with T5-XXL (11B parameters), one for factuality (which we call NLI) and one for summary quality (which we call TLDR) where we adopt the same reward model training procedure as in \citep{eisenstein2023helping}.
For our policy model, we use T5-XL (3B parameters) initialized from the FLAN checkpoint \citep{chung2024scaling} and all methods were trained for $6000$ steps. We keep other configurations unchanged from the two-reward experiments on XSum (\eg, \cref{sec:two-rewards-fixed-kl-regularizer}).
We repeat this experiment for three sampling distributions and show these results in \cref{fig:tldr-results}. We observe that \clpa and \clpall both largely Pareto-dominate the baselines while maintaining smoothly steerable behaviors Pareto-curves.
\clpl collapses to a small region likely due to its representational bottleneck.
We also observe that the steerability of the learned policy may improve as the weight sampling distribution becomes more narrow, though this may come at a cost of less Pareto-optimality.
Finally, we have observed this trend repeatedly across different model sizes, datasets and sampling distributions, which gives credence to the robustness and reliability of \clp for multi-objective finetuning.

\section{Related Works}\label{sec:related-work}
Approaches for multi-reward alignment (or MOFT) can be broadly classified into two categories: prompt-based and parameter-based conditioning. {\bf Prompt-based conditioning} approaches include Personalized Soups~\citep{jang2023personalized} which use hand-crafted prompts for personalizing LMs based on binary weights on different rewards; CPO~\citep{guo2024cpo} which employs prompting within a DPO framework; RiC~\citep{yang2024RiC}, DPA~\citep{wang2024dpa} use prompting within a supervised finetuning setup that differs from this paper which focuses on RL finetuning. On the {\bf parameter-conditioning front}, Rewarded Soups (RS) \citep{rame2023soups} presents a zero-shot approach (i.e. without multi-task training) to multi-reward alignment by inference time averaging of parameters for LMs that are independently trained to optimize each of the rewards. A more recent manuscript~\citep{zhong2024panacea} presents an approach where the reward weightings are embedded as singular values within the AdaLoRA framework~\citep{hu2021lora,zhang2023adaptive}; this can be framed as an instance of the proposed \clp framework. %

With regards to KL realignment, decoding time realignment (DeRa)~\citep{liu2024decoding} linearly mixes logits between $\piref$ and another LM learned via SOFT with the minimum KL weight $\alphamin$. \citet{shi2024decoding} showed that this idea is also effective for trading off multiple rewards.
Finally, model souping \citep{wortsman2021learning,wortsman2022model}, learning policy sub-spaces \citep{gaya2022learning,dimitriadis2023pareto}, and objective weight conditioning \citep{dosovitskiy2020you} have been applied in domains beyond LMs. We leverage these advances along with multi-task training to develop steerable LMs at inference time.

\section{Theory for Logit Mixing and CLP}\label{sec:theory}
In this section, we perform a sensitivity analysis for logit mixing and derive regret bounds for its Pareto-front.
While \clp uses parameter mixing instead of logit mixing, this analysis is still be instructive due to the similarity between parameter and ensemble mixing \citep{wortsman2022model,rame2022diverse}.

\subsection{Sensitivity Analysis for Logit Mixing}
We focus on the ``two-reward, fixed $\alpha$'' setting for simplicity and our analysis can be extended to the general case.
For any $\lambda\in[0,1]$, let $\zeta_\lambda$ be the logits of the optimal policy $\pi^\star_{\alpha,w}=\argmax_{\pi}V_{\alpha,w^\top\rvec}(\pi)$ for weightings $w=[1-\lambda,\lambda]$.
Via the analytical solution of KL-regularized reward maximization, \citet{liu2024decoding} observed that the optimal logits at $\lambda$ is expressible as mixture of the optimal logits for each individual reward, \ie, $\zeta_\lambda=(1-\lambda)\zeta_0+\lambda\zeta_1$.
Thus, given \emph{optimal} policies for $R_1$ \& $R_2$, logit-mixing provides a zero-shot way to compute the optimal policy at any intermediate $\lambda\in[0,1]$.

However, in practice, we of course are not given optimal policies and only have access to $\eps$-approximations, so it is important to understand the sensitivity of logit-mixing. We now bound the sub-optimality of logit-mixing in terms of $\eps$ and a concentrability coefficient that measures policy coverage, defined as $C_{\pi_1,\pi_2}:=\max_{x,y}\pi_2(y\mid x)/\pi_1(y\mid x)$, i.e. what is the least overlap in terms of ratio of probabilities of each policy (one per reward) over the actions (for e.g. tokens).
\begin{theorem}
Suppose $\wh\pi_1,\wh\pi_2$ are $\eps$-optimal policies for \cref{eq:single-objective-ft} with $R_1,R_2$, respectively.
For any $\lambda\in[0,1]$, let $\wh\pi_\lambda$ be the logit mixture of $\wh\pi_1$ and $\wh\pi_2$.
Then, the sub-optimality of $\wh\pi_\lambda$ is bounded by:
\begin{align*}
  \Ocal( ((1-\lambda)C_{\wh\pi_2,\wh\pi_1}^\lambda + \lambda C_{\wh\pi_1,\wh\pi_2}^{1-\lambda} + p_{\min}^{-2})\cdot\eps),
\end{align*}
where $p_{\min}$ is the minimum probability of $\wh\pi_1,\wh\pi_2$ over all input-outputs $(x,y)$ that we care about.
\end{theorem}
The coverage terms however can be infinite if the policies $\wh\pi_1,\wh\pi_2$ don't cover each other. A zero-shot method will be robust to approximations when expert policies for each individual reward cover each other. But, as we show next, zero-shot approaches will fail when this coverage condition doesn't exist anymore. The full proof is in \cref{app:theory}.

\subsection{Counterexample for zero-shot MOFT}\label{sec:counterexample-for-zero-shot}
Logit mixing cannot induce new behaviors since it can only mix behaviors from the two extremes, and so if an intermediate weighting requires a new behavior, logit mixing provably fails. Consider a toy problem with one context and three possible outputs $y_1,y_2,y_3$ with rewards $R_1(x,\cdot)=(1,0,0.75), R_2(x,\cdot)=(0,1,0.75)$. The optimal policies for $R_1,R_2$ (with $\alpha=0$) are $\pi^\star_1(x)=(1,0,0)$ and $\pi^\star_2(x)=(0,1,0)$.
However, $\pi^\star_{0.5}=(0,0,1)$, which is a \emph{qualitatively new} behavior that cannot arise from zero-shot logit mixing thus being a failure case for zero-shot logit mixing. In appendix \cref{fig:counterexample-zero-shot}, we show that the RS baseline, which is zero-shot empirically fails to learn the Pareto-optimal policy in this example, while \clp which uses multi-task training succeeds.

\section{Conclusion}
We introduced \clp, a flexible framework for MOFT that leverages techniques from multi-task training and parameter efficient finetuning to develop steerable LMs that adapt their generations to produce near Pareto optimal behavior across different weightings of individual rewards. We provide extensive benchmarking and ablations to better understand factors that enable the development of steerable LMs within the \clp framework. We supplement this with theoretical results that present conditions under which zero shot approaches work and when multi-task training is {\em necessary} to obtain near-optimal behavior. Promising future directions include (a) understanding other conditioning mechanisms such as soft tokens~\citep{hwang2023promptable}, (b) automated tuning of the weight sampling distributions~\citep{guo2019autosem}, (c) non-linear reward scalarizations~\citep{roijers2013survey} and risk-sensitive RL alignment \citep{wang2024risk}, and (d) designing adaptive algorithms with instance-dependent guarantees \citep{wang2024central}.

\section*{Acknowledgements}
We thank Jonathan Berant, Kristina Toutanova, Peter Shaw, Sertan Girgin, Jo\"elle Barral, Corinna Cortes, Slav Petrov and anonymous reviewers for helpful discussions and feedback.

\section*{Limitations}
This paper develops a framework for multi-objective finetuning and proposes variants that work out-of-the-box and are robust across different ablations, which are accompanied by auto-evaluations that present credence to these claims. We acknowledge that reward models are approximations to human interpretations of language.
When considering applications of \clp for other problem setups, it is fairly likely that additional evaluations including human evals and red-teaming need to be considered to mitigate any risks posed by a more flexible LM. The specific issues that one would encounter when pursuing extensions along these directions is beyond the scope of this paper.

\section*{Ethical Considerations}
This paper presents finetuning techniques for learning more flexible LMs that can provide generations that trade-off multiple potentially conflicting objectives at inference time. The design of objectives for alignment is an active area of research in itself. By having an LM that can adapt to provide optimized generations for different weighting of objectives, one can increase the risk of having LMs exhibit behaviors contrary to societal norms and values, and so must be subject to vetting, red-teaming and other protocols to ensure these models don't fall foul of societal norms.

\bibliographystyle{plainnat}
\bibliography{main}

\newpage
\appendix
\onecolumn
\begin{center}\LARGE
\textbf{Appendices}
\end{center}

\section{Additional Algorithmic Details}
\subsection{Illustration of CLP}
The following supplemental figure illustrates one training round of \clp (\cref{alg:CLP-framework}).
\begin{figure}[h!]
  \centering
  \includegraphics[width=0.5\linewidth]{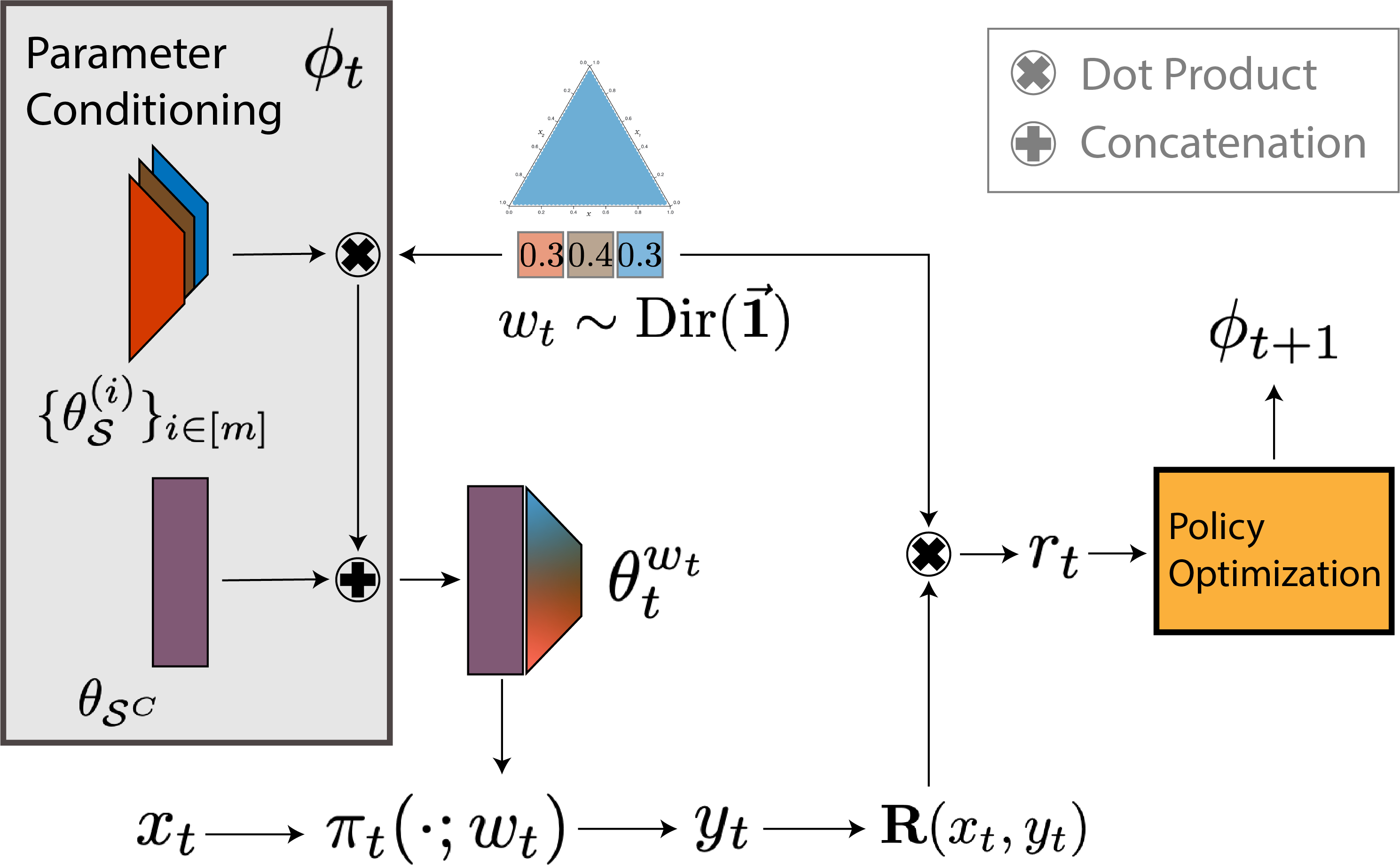}
  \caption{\clp samples diverse reward weightings $w_t$ at training time to facilitate multi-task learning and uses parameter-space conditioning to compute a conditioned policy $\pi_t(\cdot;w_t)$.
  }
  \label{fig:clp}
\end{figure}

\subsection{Prompt-based Conditioning}\label{app:prompt-based}
We use prompt-conditioning only to trade-off between rewards (\ie, not for trading off KL).
Given an input prompt $x$ and reward weightings $w$, we propose to concatenate $w$ to the prompt by prefixing $x$ with a formatted string.
For example if $R_1$ is NLI (a measure of factuality) and $R_2$ is Rouge, and $w=[0.4, 0.6]$, we used the the prefix string
\begin{equation*}
    p = \texttt{``factuality: 0.4 0.4 0.4 0.4 0.4; rouge: 0.6 0.6 0.6 0.6 0.6.''}.
\end{equation*}
We repeat the weights multiple times such that the LM can better pay attention to it and we found $5$ repetitions to be reasonable.
We found that repeating weights multiple times is helpful but the benefits are not monotonic with more repetitions.

\subsection{Analysis of CLP Gradients}\label{app:gradient-analysis}
While computing \clp's gradient $g_t$ in \cref{line:update-phi-parameters} is easy with autodiff libraries,
it is insightful to analyze how gradient propagates through \clp's parameter conditioning step.
Let $\gamma_t = r_t \nabla_{\theta_t^{\alpha_t,w_t}}\log\pi_t(y_t\mid x_t;\alpha_t,w_t)$ denote the gradient w.r.t. LM parameters evaluated at $\theta_t^{\alpha_t,w_t} = \theta_{t,\Scal^C}\oplus\theta_{t,\Scal}^{\alpha_t,w_t}$, where $\theta_{t,\Scal^C}$ is the unconditioned parameter and $\theta^{\alpha_t,w_t}_{t,\Scal}$ is the $(\alpha_t,w_t$)-conditioned $\Scal$-parameter at round $t$.
We decompose $g_t=(g_{t,\Scal^C},\{g_{t,\Scal}^{(i)}\}_{i\in[m]})$ where $g_{t,\Scal^C} = \gamma_t[\Scal^C]$, and $g_{t,\Scal}^{(i)} = (1-\mixmap(\alpha_t))\cdot w_t[i] \cdot \gamma_t[\Scal]$ for all $i\in[m]$.
Thus, the gradient for \clp's unconditioned parameters is exactly the LM gradient at $\Scal^C$, and the gradient for \clp's conditioned parameters point in the direction of the LM gradient at $\Scal$, and are scaled by the reward weights $w_t$ and $1-\mixmap(\alpha_t)$.
\Eg, $i$-th conditioned parameter $\theta_{t,\Scal}^{(i)}$ is updated more if $R_i$ has high weight (\ie, $w_t[i]$ is large); conversely, it is not updated at all if $R_i$ has zero weight.
The scaling by $1-\mixmap(\alpha_t)$ implies that the conditioning parameters are updated more when $\alpha_t$ is small and close to $\alphamin$, and updated less when $\alpha_t$ is close to $1$.

\subsection{Weightings Distribution and KL-Mixing}\label{sec:clp-sampling}
We suggest a default sampling distribution $\Qcal$ for the reward and KL weightings. We also suggest a compatible KL-mixing function $\mixmap$, which transforms $\alpha$ before applying parameter-mixing in the conditioning mechanism.

First, a natural sampling distribution for reward weights $w\in\Delta_m$ is the Dirichlet distribution $\Qcal_w=\op{Dir}(\bm{\beta})$ with parameters $\bm{\beta}\in\RR^m_+$.
Concentrated sampling (\eg, large $\bm{\beta}[i]\to\infty$ values) can more easily cause unsteerable ``mode-collapsed'' behavior. %
From our experience, a good default is uniform Dirichlet with $\bm{\beta}=(1,1,\dots,1)$ which consistently led to steerable behaviors in \clpall and \clpa.

Next, we discuss how to set the KL-mixer map $\mixmap(\alpha)$.
As motivation, recall the following theorem about KL-realignment from \citet{liu2024decoding}.
Let $\zeta_\alpha(x)\in\RR^{|\Ycal|}$ be the logits of the optimal policy with KL-weight $\alpha$.
Then, $\zeta_\alpha(x) = \zetaref(x) + (1-\alpha) R(x,\cdot)/\alpha$, where $\zetaref$ are the logits of $\piref(x)$.
Rearranging terms, for all $\alpha\in[\alphamin,1]$, we have $\zeta_{\alpha}=(1-\beta)\zeta_{\alphamin}+\beta\cdot\zetaref$ where $\beta=\mixmap(\alpha)$ and
\begin{equation}
  \mixmap(\alpha)=\frac{\alpha-\alphamin}{\alpha(1-\alphamin)}. \label{eq:softmdp-mixing-function}
\end{equation}
Note that $\mixmap$ satisfies $\mixmap(1)=1$, so when $\alpha=1$ we use purely $\thetaref[\Scal]$ and do not mix in any learned $\theta_\Scal$; this makes intuitive as the minimizer of KL is indeed $\piref$. Conversely, $\mixmap(\alphamin)=0$, meaning that $\thetaref$ is not mixed in when $\alpha=\alphamin$.
Thus, since $\mixmap$ is the right mixing function for logits, we adopt this $\mixmap$ for general parameter mixing in \clp.

To sample KL weights $\alpha_t\sim\Qcal_\alpha$, we suggest to use the inverse CDF method with $\mixmap^{-1}(u) = \frac{\alphamin}{\alphamin \cdot u + (1-u)}$, \ie, $\alpha_t=\mixmap^{-1}(U)$ with $U\sim\op{unif}[0,1]$.
This ensures that the KL-mixing weights are uniformly distributed, \ie, $\mixmap(\alpha_t)\sim\op{unif}[0,1]$. %
Since the median is $\frac{2\alphamin}{\alphamin+1}$, this distribution also places higher probabilities on smaller $\alpha$ values.
This is desirable since small $\alpha$ corresponds to larger learning signals (large $\alpha$ just forces the model to be close to $\thetaref$).
In \cref{app:kl-weight-distribution-and-mixing-function}, we show samples from $\alpha_t\sim\Qcal_\alpha$ and $\mixmap(\alpha_t)$.
Finally, we note that $w$ or $\alpha$ can be fixed to a specific value if one cares about only varying the other.

\subsection{KL weight distribution and mixing function}\label{app:kl-weight-distribution-and-mixing-function}
\begin{figure}[h!]
\begin{subfigure}{0.4\textwidth}
\includegraphics[width=\linewidth]{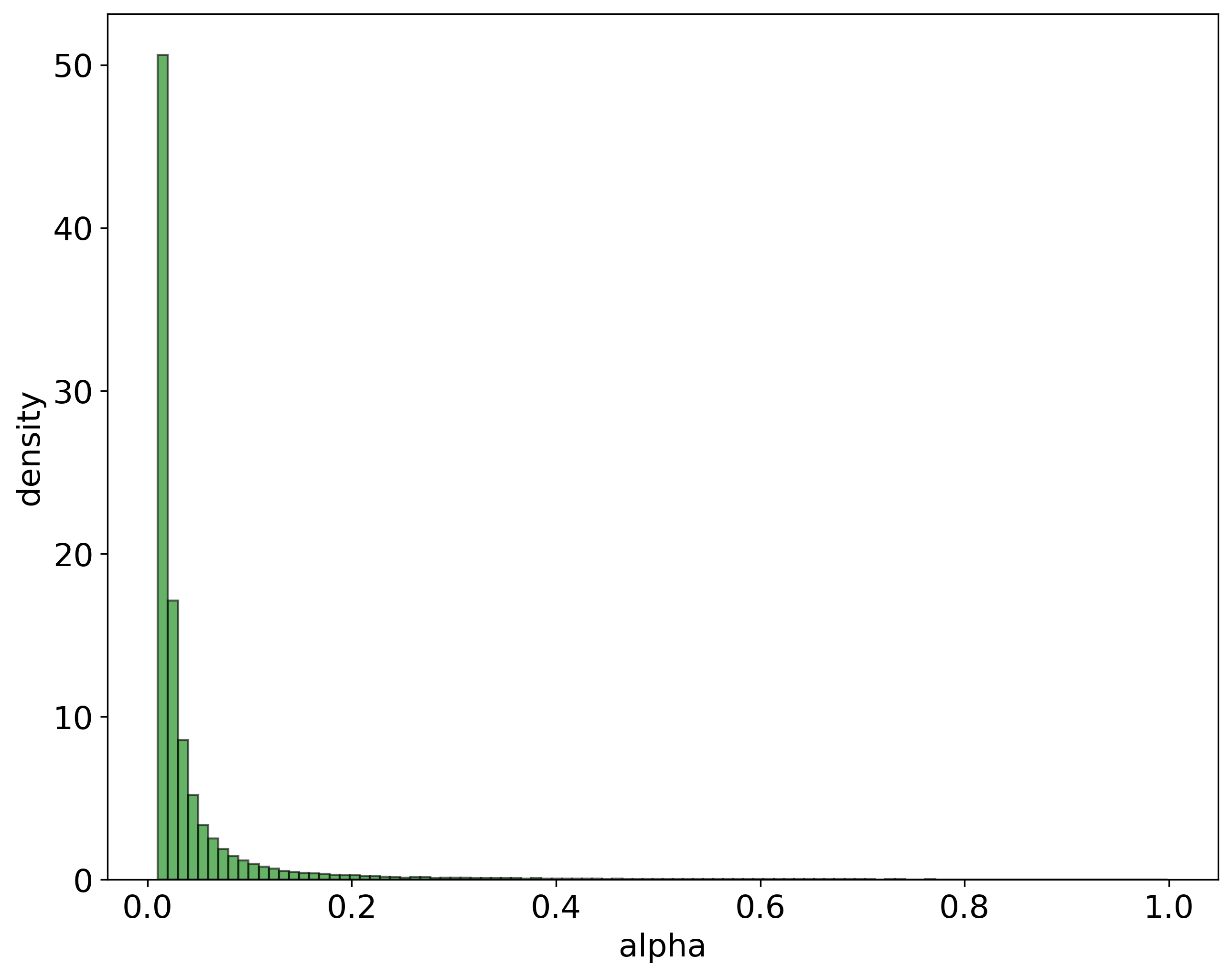}
\caption{Distribution of $\alpha_t\sim\Qcal_\alpha$.}
\end{subfigure}\hspace*{\fill}
\begin{subfigure}{0.4\textwidth}
\includegraphics[width=\linewidth]{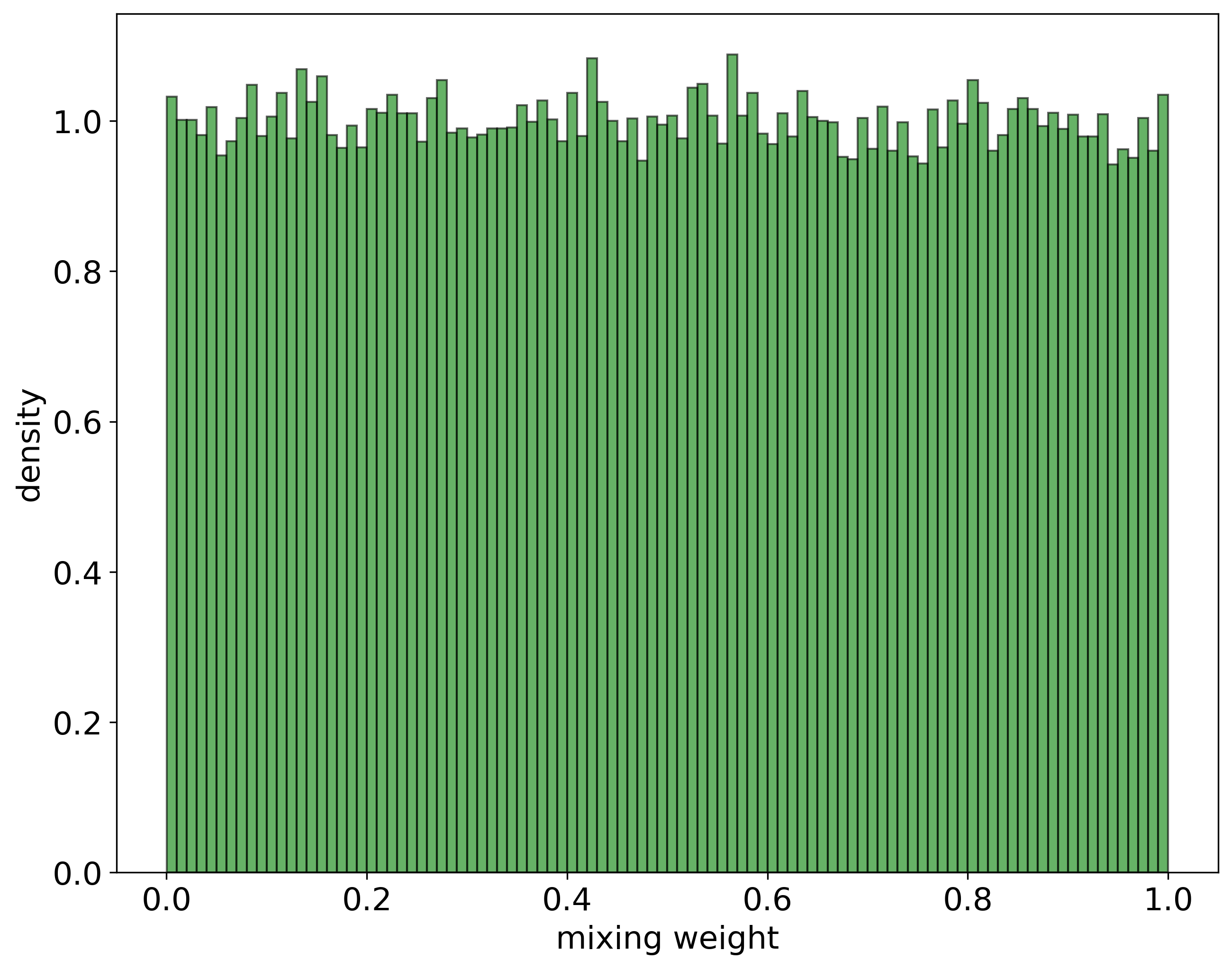}
\caption{distribution of $\mixmap(\alpha_t)\sim\op{unif}[0,1]$.}
\end{subfigure}\hspace*{\fill}
\caption{Distributions of $\alpha_t\sim\Qcal_\alpha$ and $\mixmap(\alpha_t)$ with $\Qcal_\alpha,\mixmap$ defined in \cref{sec:clp-sampling}.} \label{fig:inverse-sampling-and-softmdp-mixing}
\end{figure}

\section{Experiment Details}\label{app:more-training-details}
\noindent\textbf{Reward and normalization.}
For Rouge, we use the LSum variant \citep{lin2004rouge} throughout the paper. Both the NLI and TLDR models were trained with T5-large \citep{2020t5}.
We normalize rewards such that they all approximately lie in the range $[0,1]$ to ensure they are on a comparable scale. This is important as otherwise it is easy for a reward with large scale to dominate the objective. In particular, we linearly mapped the following ranges for each reward to $[0,1]$: $\{\RNLI: (-7, 0), \RRouge: (19, 29), \RTLDR: (-1.6, 2)\}$.

\noindent\textbf{KL regularizer.}
In the single-reward, multi-KL experiments, we use $\alpha_{\min}=0.01$, and in the two-reward and three-reward experiments with fixed KL regularizer, we use $\alpha=0.01$.
This $\alpha$ value is chosen based on observing that, under our normalized reward setup, it is high enough to prevent reward hacking and low enough to exhibit interesting qualitative differences from the SFT $\piref$.

\noindent\textbf{Training.}
To train \clp and all our baselines, we use a batch size of $32$.
For \clp, we sample fresh reward or KL weightings per batch, instead of per example, as it is more efficient to only condition once per batch. We also do this for \clpp for consistency. We use the T5X implementation \citep{roberts2023scaling} and the Adafactor optimizer \citep{shazeer2018adafactor}.
In terms of training budget, we ran $10,000$ iterations for the single-reward, multi-KL experiments, and we ran $30,000$ iterations for the multi-reward, fixed-KL experiments. For the RS baseline, the training steps were divided evenly between the rewards; \eg, in the two-reward setting, RS learns two LMs each trained for $15,000$ and initialized from $\piref$.
Thus \clp and RS are trained with the same number of iterations.
For all results, we train on the XSum training set and report the reward functions and/or qualitative generations on the \emph{validation set}.

\subsection{Policy Optimization}
Recall that the policy optimization step in \cref{line:update-phi-parameters} can be implemented with any RL method (policy gradients, actor-critic etc.).
This paper uses REINFORCE with a control variate (i.e., baseline) which uses fewer hyper-parameters and lighter implementation than PPO \citep{schulman2017proximal}, and has been successfully used in prior works  \citep{roit2023factually,huang2024rlhf}.
To train the control variate, we maintain a value network of the same architecture as the policy network where the prediction is the logit value for some fixed token id.
For example, in a problem with $m$ rewards, we use the first $m$ token ids where the $i$-th token's logit serves as the value function for the $i$-th reward, \ie, $\wh V_i(x)\approx\EE_{w\sim\Qcal,y\sim\pi(\cdot;w)}R_i(x,y)$. This value network is trained by minimizing MSE, which is standard in RL. Then, these predictions are linearly combined with weightings to compute the value estimate for the weighted reward, \ie, $\wh V_t = \sum_i w_i\cdot \wh V_i(x_t)$.
The advantage is computed as usual $A_t = r_t-\wh V_t$ and is batch-normalized before being multiplied with the $\nabla_{\phi_t}\log\pi_t$ term.

\begin{center}
\begin{tabular}{ |p{5cm}|p{9.5cm}|  }
\hline
Hyper-Parameter & Value\\ \hline
\multicolumn{2}{|c|}{{\bf Model Family Details}} \\
\hline
Policy/Value Model & \makecell{T5-Base (220M) \\(ablations: T5-small (60M) and T5-large (770M))}\\\hline
Reward Models & \makecell{Rouge LSum\\ NLI/TLDR (learnt - T5-large)}\\\hline
Reward Normalization & \makecell{$\RNLI: (-7,0)\rightarrow(0,1)$\\$\RTLDR: (-1.6, 2)\rightarrow(0,1)$\\$\RRouge: (19, 29)\rightarrow(0,1)$}\\\hline
Tokenizer & Sentence Piece Tokenizer (32k vocabulary size) \\\hline
Code & T5X \citep{roberts2023scaling} \\\hline
Computing Infra & TPU-v5e chips \\\hline
Experiment Time & Using $16$ TPU-v5e chips, our T5-base runs that includes $30k$ training steps and evaluations every $10k$ steps took $8$ hours. Note that each evaluation over the validation set takes roughly $1$ hour. \\
\hline
\multicolumn{2}{|c|}{{\bf Policy Optimization Hyper-parameters}} \\
\hline
Batch Size & \makecell{32} \\\hline
Policy Optimizer & \makecell{Adafactor~\citep{shazeer2018adafactor}\\ learning rate $3e-5$} \\\hline
Value Optimizer & \makecell{Adafactor~\citep{shazeer2018adafactor}\\ learning rate $1e-4$}\\\hline
\multicolumn{2}{|c|}{{\bf \clp/Multi-task training Hyper-parameters}} \\
\hline
KL-Strength $\alpha$&\makecell{\underline{Multi-Reward Single KL}: $0.01$\\Also tried $0.1$ in ablations\\\underline{Single-Reward Multi KL}: $\alpha_{\min}=0.01$}\\\hline
Training Budget & \makecell{\underline{2 Reward, Single KL}: \\ Rewarded Soups: $15,000$ ($\times 2$, one per reward)\\\clp variants: $30,000$.\\\underline{Single Reward, Multi KL}:TODO:FILL\\\underline{3 Reward, Single KL}:\\Rewarded Soups: $10,000$ ($\times 3$, one per reward)\\\clp variants: $30,000$.}\\\hline
Sampling Distribution $\Dcal$ & \makecell{Dirichlet(1.0) for main results\\Also Dirichlet(0.3) in ablations.}\\\hline
\end{tabular}
\end{center}

\newpage
\section{Additional Experimental Results}

\subsection{Additional Multi-KL Results}\label{app:multi-kl-rewards}
This section focuses on the single-reward, multi-KL setup as in \cref{sec:single-reward-varying-kl}. We report \clp results with two choices of the KL-mixing function $\mixmap$.
First, linear mixing function is $\mixmap(x)=x$.
Second, softmdp mixing function is defined in \cref{eq:softmdp-mixing-function} of \cref{sec:clp-sampling}, which is more principled from the perspective of DeRa \citep{liu2024decoding}. We see that softmdp mixing indeed yields better steerability for \clp.

\begin{figure}[!h]
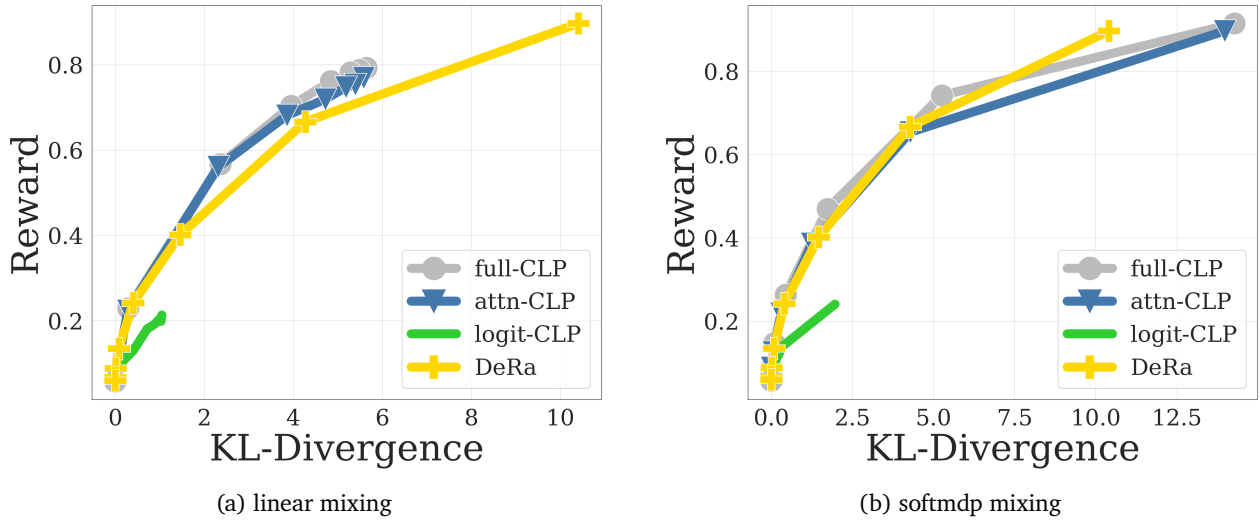

\begin{subfigure}{0.48\textwidth}
\includegraphics[width=\linewidth]{without_so_plots/so_linear_reward_v_kl_alpha_0_01.pdf}
\caption{linear mixing}
\end{subfigure}\hspace*{\fill}
\begin{subfigure}{0.48\textwidth}
\includegraphics[width=\linewidth]{without_so_plots/so_softmdp_reward_v_kl_alpha_0_01.pdf}
\caption{softmdp mixing}
\end{subfigure}
\caption{Ablation on the KL-weight mixing function $\mixmap$. Linear mixing sets $\mixmap(\alpha)=\alpha$. Softmdp mixing is the one described in \cref{sec:clp-sampling}.}
\end{figure}

\newpage
\subsection{Additional Two-Reward Results}\label{app:base-two-reward-fixed-alpha}
Results for (1) NLI v. Rouge, (2) NLI v. TLDR for fixed $\alpha\in\{0.01, 0.1\}$.

\begin{figure}[h!]
\centering
\begin{subfigure}{0.48\textwidth}
\includegraphics[width=\linewidth]{without_so_plots/abridged_base_nli_rouge_alpha_0_01.pdf}
\caption{$\alpha=0.01$, nli vs. rouge} \label{fig:nli-rouge-1e-2}
\end{subfigure}
\begin{subfigure}{0.48\textwidth}
\includegraphics[width=\linewidth]{without_so_plots/abridged_base_nli_tldr_alpha_0_01.pdf}
\caption{$\alpha=0.01$, nli vs. tldr} \label{fig:nli-tldr-1e-2}
\end{subfigure}\hspace*{\fill}

\medskip
\begin{subfigure}{0.48\textwidth}
\includegraphics[width=\linewidth]{without_so_plots/abridged_base_nli_rouge_alpha_0_1.pdf}
\caption{$\alpha=0.1$, nli vs. rouge} \label{fig:nli-rouge-1e-1}
\end{subfigure}
\begin{subfigure}{0.48\textwidth}
\includegraphics[width=\linewidth]{without_so_plots/base_nli_tldr_alpha_0_1.pdf}
\caption{$\alpha=0.1$, nli vs. tldr} \label{fig:nli-tldr-1e-1}
\end{subfigure}\hspace*{\fill}

\caption{Plots comparing \clp instances (\clpall, \clpa, \clpl) against Rewarded Soups (RS)~\citep{rame2023soups} and \clpp in the two-reward experiments.} \label{fig:two-reward-comparisons}
\end{figure}

\newpage
\subsection{Additional Three-Reward Results}\label{app:base-three-reward-fixed-alpha}

\begin{figure}[!h]
\begin{subfigure}{0.48\textwidth}
\includegraphics[width=\linewidth]{without_so_plots/full_attn_clp_prompt_rs_three_rewards_alpha_0_01.png}
\caption{\clpall, \clpa, \clpp}
\end{subfigure}\hspace*{\fill}
\begin{subfigure}{0.48\textwidth}
\includegraphics[width=\linewidth]{without_so_plots/logit_prompt_rs_three_rewards_alpha_0_01.png}
\caption{\clpl, \clpp}
\end{subfigure}
\caption{Plots comparing \clp instances (\clpall, \clpa, \clpl) against Rewarded Soups (RS)~\citep{rame2023soups} and \clpp in the three-reward experiments.}
\end{figure}

\subsection{Effect of Training Iterations}\label{app:ablation-training-iterations}
\begin{figure}[!h]
\includegraphics[width=\linewidth]{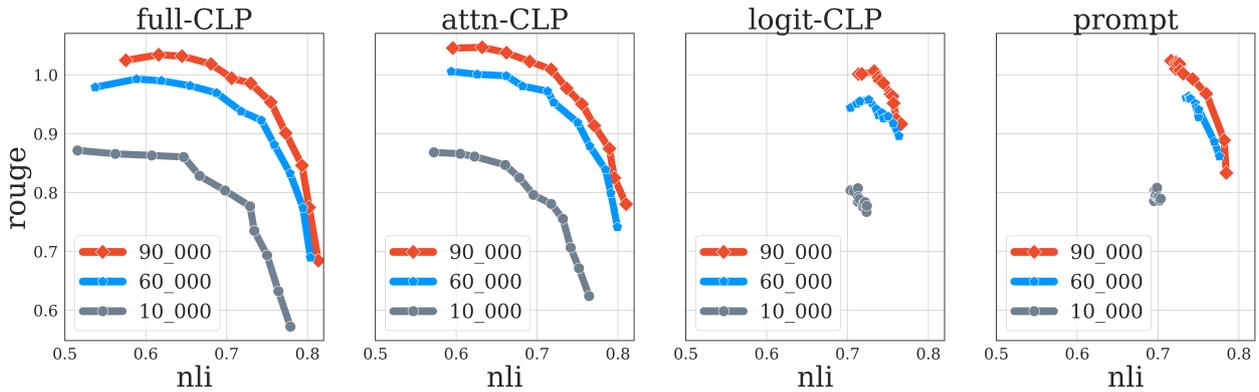}
\caption{Effect of Training Iterations. Observe that \clp variants such as \clpall and \clpa tend to present reasonably strong spread out behaviors even with smaller training budgets. \clpp and \clpl tend to start spreading out just a little after running the expensive RLHF training procedure for $3\times$ -- note that the spread of these methods is still far inferior to ones obtained by \clpall and \clpa.}
\end{figure}

\newpage
\subsection{Combining Prompt and Parameter Conditioning}\label{app:combining-prompt-parameter-conditioning}

\begin{figure}[h!]
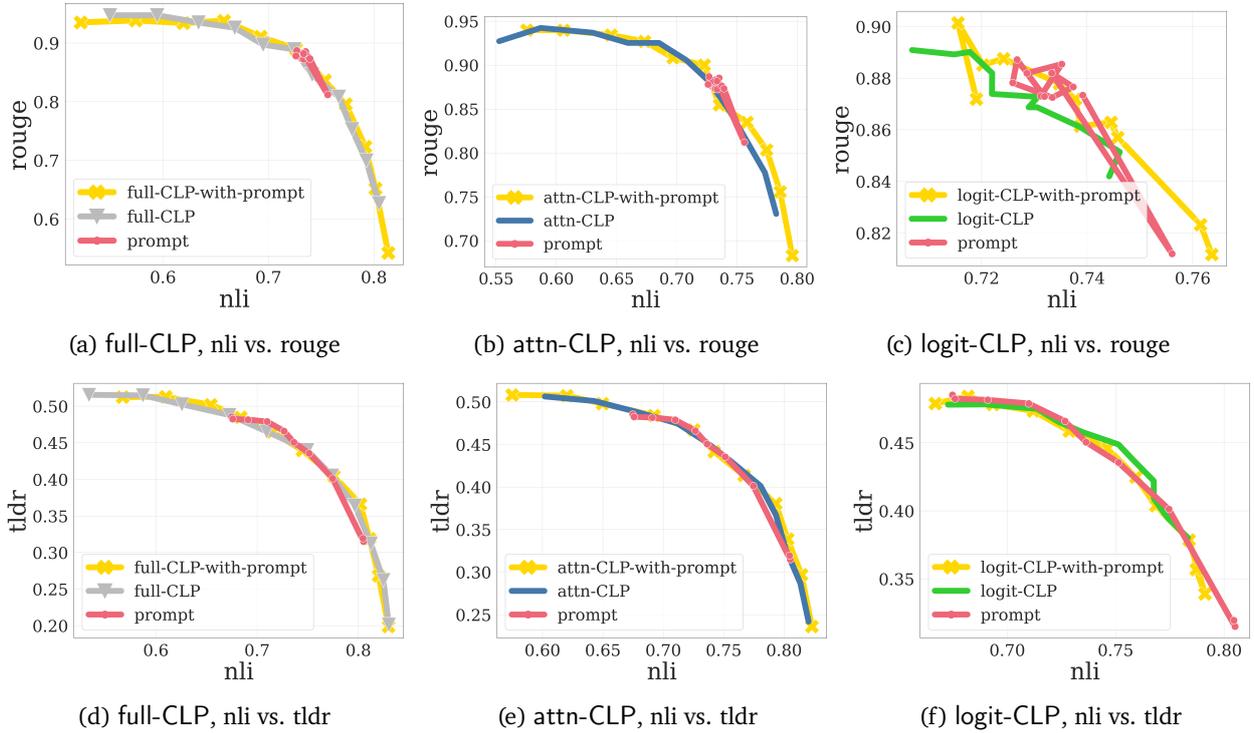

\begin{subfigure}{0.32\textwidth}
\includegraphics[width=\linewidth]{prompt_ablation_and_longrun/full_CLP_base_nli_rouge_alpha_0_01.pdf}
\caption{\clpall, nli vs. rouge}
\end{subfigure}
\begin{subfigure}{0.32\textwidth}
\includegraphics[width=\linewidth]{prompt_ablation_and_longrun/attn_CLP_base_nli_rouge_alpha_0_01.pdf}
\caption{\clpa, nli vs. rouge}\label{fig:combine-prompt-attn-nli-rouge}
\end{subfigure}
\begin{subfigure}{0.32\textwidth}
\includegraphics[width=\linewidth]{prompt_ablation_and_longrun/logit_CLP_base_nli_rouge_alpha_0_01.pdf}
\caption{\clpl, nli vs. rouge}\label{fig:combine-prompt-logit-nli-rouge}
\end{subfigure}

\medskip
\begin{subfigure}{0.32\textwidth}
\includegraphics[width=\linewidth]{prompt_ablation_and_longrun/full_CLP_base_nli_tldr_alpha_0_01.pdf}
\caption{\clpall, nli vs. tldr}
\end{subfigure}\hspace*{\fill}
\begin{subfigure}{0.32\textwidth}
\includegraphics[width=\linewidth]{prompt_ablation_and_longrun/attn_CLP_base_nli_tldr_alpha_0_01.pdf}
\caption{\clpa, nli vs. tldr}
\end{subfigure}\hspace*{\fill}
\begin{subfigure}{0.32\textwidth}
\includegraphics[width=\linewidth]{prompt_ablation_and_longrun/logit_CLP_base_nli_tldr_alpha_0_01.pdf}
\caption{\clpl, nli vs. tldr}
\end{subfigure}\hspace*{\fill}
\caption{Ablation study that enables prompting for different \clp instances, across different experiments with a pair of reward functions.} \label{fig:prompt-parameter-conditioning-ablation-appendix}
\end{figure}
In the above figures, we see that prompt-conditioning does not always contribute to more steerability in \clp, but it also does not hurt the steerability. We remark that the aberration in \cref{fig:combine-prompt-logit-nli-rouge} is simply a zoomed in version of \cref{fig:combine-prompt-attn-nli-rouge}, showing that \clpl and \clpp are not very steerable in this example.

\newpage
\subsection{Ablation on Model Size}\label{app:ablation-model-size}
We present results across two paired-reward setups (NLI vs. Rouge, NLI vs. TLDR) and three model sizes (T5-small, T5-base, T5-large).
In T5-large which has $24$ attention layers, we only mix the first $12$ layers of the attention parameters for \clpa.
We believe that mixing all attention layers can only further improve steerability. For the T5-large model, the first $12$ attention layer parameters account for around $20\%$ of the total parameter count.

\begin{figure}[h!]
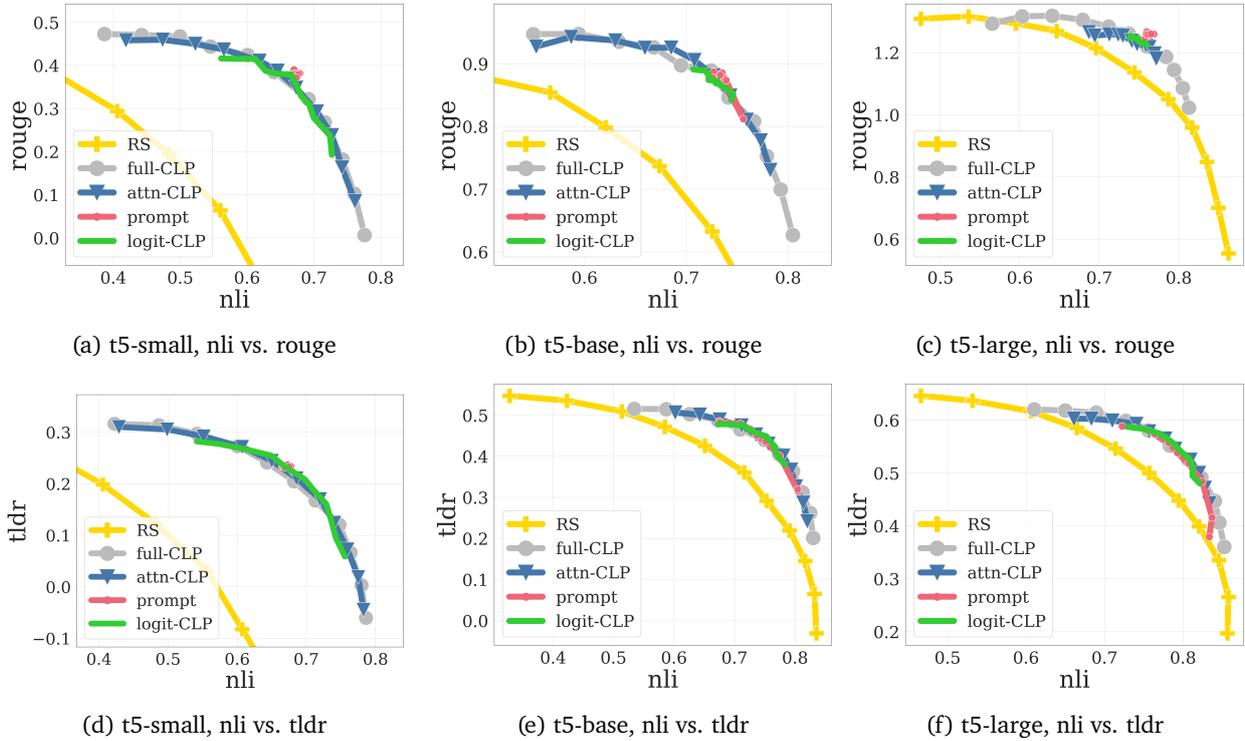


\begin{subfigure}{0.32\textwidth}
\includegraphics[width=\linewidth]{without_so_plots/abridged_small_nli_rouge_alpha_0_01.pdf}
\caption{t5-small, nli vs. rouge}
\end{subfigure}\hspace*{\fill}
\begin{subfigure}{0.32\textwidth}
\includegraphics[width=\linewidth]{without_so_plots/abridged_base_nli_rouge_alpha_0_01.pdf}
\caption{t5-base, nli vs. rouge}
\end{subfigure}
\begin{subfigure}{0.32\textwidth}
\includegraphics[width=\linewidth]{without_so_plots/large_nli_rouge_alpha_0_01.pdf}
\caption{t5-large, nli vs. rouge}
\end{subfigure}\hspace*{\fill}

\medskip
\begin{subfigure}{0.32\textwidth}
\includegraphics[width=\linewidth]{without_so_plots/abridged_small_nli_tldr_alpha_0_01.pdf}
\caption{t5-small, nli vs. tldr}
\end{subfigure}\hspace*{\fill}
\begin{subfigure}{0.32\textwidth}
\includegraphics[width=\linewidth]{without_so_plots/base_nli_tldr_alpha_0_01.pdf}
\caption{t5-base, nli vs. tldr}
\end{subfigure}
\begin{subfigure}{0.32\textwidth}
\includegraphics[width=\linewidth]{without_so_plots/large_nli_tldr_alpha_0_01.pdf}
\caption{t5-large, nli vs. tldr}
\end{subfigure}\hspace*{\fill}
\caption{Ablation study involving different model sizes for parameterizing the steerable policy with the t5 family of models. Observe that \clp variants tend to perform in a predictable and robust manner across these model sizes and choices of rewards. Other distinct observations include (a) \clpp behaves in rather unpredictable manners (collapsing and spreading out in rather non-transparent manners, see nli vs. rouge plots in 2nd row), (b) Rewarded Soups show predictable behavior that attempts to catch up with \clp variants but still retaining a healthy gap in terms of Pareto front even with t5-large model sizes.} \label{fig:model-size-ablation-appendix}
\end{figure}

\newpage
\subsection{Ablation on Sampling Strategy}\label{app:sampling-strategy-ablation}

\begin{figure}[h!]
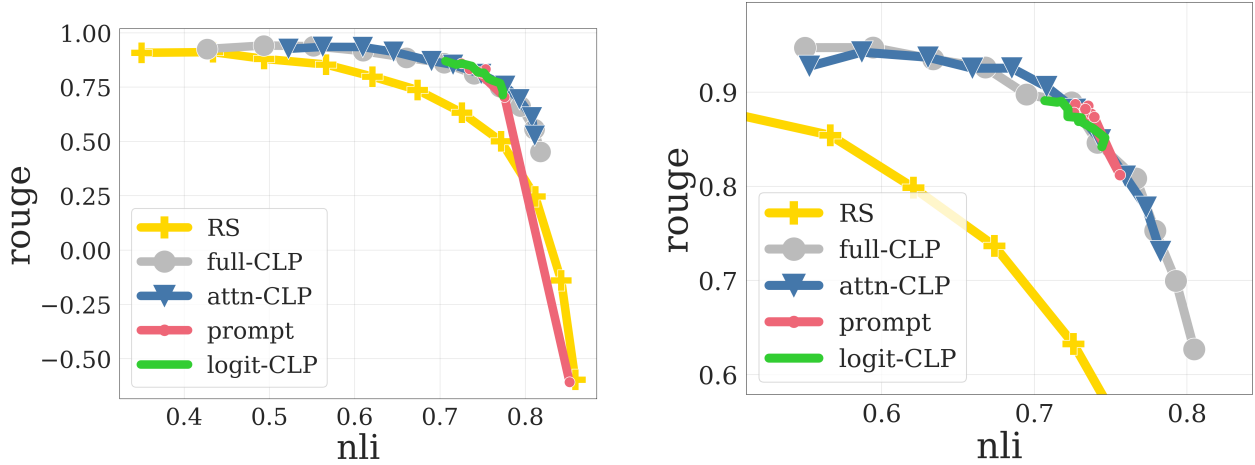

\begin{subfigure}{0.48\textwidth}
\includegraphics[width=\linewidth]{without_so_plots/dirichlet_03_base_nli_rouge_alpha_0_01.pdf}
\caption{nli vs. rouge, sampling$\sim~\textrm{Dirichlet}(0.3)$}
\end{subfigure}\hspace*{\fill}
\begin{subfigure}{0.48\textwidth}
\includegraphics[width=\linewidth]{without_so_plots/abridged_base_nli_rouge_alpha_0_01.pdf}
\caption{nli vs. rouge, sampling$\sim~\textrm{Dirichlet}(1.0)$}
\end{subfigure}
\caption{Ablation on the weight sampling distribution. Setting is NLI v. Rouge with $\alpha=0.01$. Observe that the behaviors of \clp variants tend to offer broadly consistent behaviors regardless of the choice of the reward weighting sampling distributions $\Dcal$ while \clpp tends to be sensitive to this choice -- finding an reasonable sampling strategy for a new problem can be tricky to solve when employing \clpp, while \clp variants remain robust to this choice.} \label{fig:sampling-strategy-ablation}
\end{figure}

\newpage
\section{Additional Qualitative Results}\label{app:generations}
\subsection{Details of Automatic Evaluation}\label{app:details-autorater}
We evaluate generations from \clp variants (\clpall, \clpa, \clpl) against both \clpp and Rewarded Soups baselines. For this, we pick the NLI vs. TLDR setup, use the T5-large models trained in \cref{sec:ablation-model-size}) and evaluate on $2000$ prompts from the XSum validation set.

We employ the following prompt to compare which one of a pair of summaries is more concise while being reasonable in quality in terms of capturing relevant information in the article.
\begin{tcolorbox}
You are an expert summary rater. Given an ARTICLE and two summaries SUMMARY1 and SUMMARY2, compare SUMMARY1 and SUMMARY2 based on how **concise the summary is in capturing relevant information** in the ARTICLE. \\\\ Output SUMMARY1 if it **more concisely captures relevant information** in the ARTICLE compared to SUMMARY2. Alternatively, output SUMMARY2 if it **more concisely captures relevant information** in the ARTICLE compared to SUMMARY1.\\\\ ARTICLE: <Insert article to summarize here.> \\ SUMMARY1: <Insert summary from ALG1 here.> \\ SUMMARY2: <Insert summary from ALG2 here.> \\ Which summary **more concisely captures information** in the ARTICLE? You can answer one of SUMMARY1 or SUMMARY2. ANSWER:
\end{tcolorbox}
Similarly, for capturing how comprehensive and high quality a summary is relative to the other, we employ the following prompt.
\begin{tcolorbox}
You are an expert summary rater. Given an ARTICLE and two summaries SUMMARY1 and SUMMARY2, compare SUMMARY1 and SUMMARY2 based on **the quality and comprehensiveness of the summary** in the ARTICLE. \\\\ Output SUMMARY1 if it is of **higher quality and more comprehensive** in summarizing the ARTICLE compared to SUMMARY2. Alternatively, output SUMMARY2 if it is of **higher quality and more comprehensive** in summarizing the ARTICLE compared to SUMMARY1.\\\\ ARTICLE: <Insert article to summarize here.> \\ SUMMARY1: <Insert summary from ALG1 here.> \\ SUMMARY2: <Insert summary from ALG2 here.> \\ Which summary is of **higher quality and more comprehensive** in capturing information in the ARTICLE? You can answer one of SUMMARY1 or SUMMARY2. ANSWER:
\end{tcolorbox}

\noindent For each instance in the validation set, we compare swapped pairs (SUMMARY1, SUMMARY2) and (SUMMARY2, SUMMARY1) with an automatic evaluator~\citep{team2023gemini} for both the conciseness prompt and comprehensiveness prompt. We consider it to be a win for one of the algorithm if it wins with both of the swapped pairs on one of the prompts while at least obtaining a tied result on the other prompt.

\subsection{Generations for NLI v. Rouge}
In this section, we provide sample summaries from \clpall, \clpa, \clpl and the \clpp baseline on NLI v. Rouge where we evaluate three weightings $w_1=(0.8,0.2),w_2=(0.5,0.5),w_3=(0.2,0.8)$.

\newcommand{\sizeOfAlgorithmCol}{1.5cm}
\newcommand{\sizeOfSummaryCol}{4.1cm}
\newcommand{\sizeOfLongTable}{13.8cm}
\subsubsection{Document 9}
\begin{longtable}{|p{\sizeOfAlgorithmCol}|p{\sizeOfSummaryCol}|p{\sizeOfSummaryCol}|p{\sizeOfSummaryCol}|}
  \hline
  \multicolumn{4}{|p{\sizeOfLongTable}|}{\textbf{Document \#9}. summarize: Cash, who still presents shows on BBC radio in the South, played his first record, Bill Haley and His Comets' Rock Around the Clock, in Canada. "This little job came up as holiday relief on 15 August 1964," he said. "It feels more like 50 minutes than 50 years. There were times when things weren't the best in the west, but 99.99\% of the time they were great." Cash, aged 72, was born in London but went to Canada with the Merchant Navy. His first broadcast give him an audition tape to take to the pirate ships broadcasting offshore to Britain, and he joined Radio London. "I saw this guy huddling over and being very ill, because the ship was going up and down at a heck of a pace," he said. "He said, 'I'm Maurice Cole - they call me Kenny Everett'." When the pirates, which inspired the 2009 film The Boat That Rocked, were outlawed, Cash became one of the first DJs on Radio 1 in 1967 alongside the likes of Tony Blackburn, Pete Murray, Ed Stewart and John Peel. He also worked for Capital Radio for 20 years from 1973 before rejoining the BBC in 1999. "I love talking to the people and getting their reaction," he said. "And, of course, you get to play the best music on the planet, which can't be bad." Cash's shows are broadcast at weekends in Kent - where he lives - Surrey, Sussex, Berkshire, Dorset, Oxfordshire and Hampshire. There will be a special show on 25 August, but he said he was spending the 50th anniversary of his first show getting an MOT for his car. "Maybe I'll have a cup of coffee or something later on with the wife." \newline \textbf{Ground Truth Summary: }Former pirate radio and Radio One DJ Dave Cash is celebrating 50 years of broadcasting. } \\
  \hline
  Algorithm & High nli, Low rouge & Mid nli, Mid rouge & Low nli, High rouge \\
  \hline
  full-CLP & A radio DJ has marked the 50th anniversary of his first broadcast. & A BBC DJ has marked the 50th anniversary of his first broadcast on the air. & DJ Kenny Cash has marked the 50th anniversary of his first radio show on Radio London 50 years ago. \\
  \hline
  attn-CLP & DJ Kenny Cash has marked the 50th anniversary of his first radio show. & DJ Kenny Cash has marked the 50th anniversary of his first radio show. & DJ Kenny Cash has marked the 50th anniversary of his first radio show 50 years ago. \\
  \hline
  logit-CLP & Kenny Cash has celebrated the 50th anniversary of his first radio show. & Kenny Cash has celebrated the 50th anniversary of his first radio show. & Kenny Cash has celebrated the 50th anniversary of his first radio show. \\
  \hline
  prompting & DJ Kenny Cash has marked the 50th anniversary of his first broadcast on the BBC. & DJ Kenny Cash has marked the 50th anniversary of his first radio broadcast. & DJ Kenny Cash has marked the 50th anniversary of his first broadcast on the BBC. \\
  \hline
\end{longtable}

\subsubsection{Document 11}
\begin{longtable}{|p{\sizeOfAlgorithmCol}|p{\sizeOfSummaryCol}|p{\sizeOfSummaryCol}|p{\sizeOfSummaryCol}|}
  \hline
  \multicolumn{4}{|p{\sizeOfLongTable}|}{\textbf{Document \#11}. summarize: The game set a new record for the most-watched event in US television, beating last year's audience of 112.2 million. Viewing peaked at 120.8 million during the last quarter, which saw the New England Patriots beat the Seattle Seahawks 28-24. Some 118.5 million watched Katy Perry's half-time show - three million more than last year's show with Bruno Mars. Ratings compiler Nielsen said the 12-minute performance - which also featured Lenny Kravitz and Missy Elliot - attracted the biggest half-time audience since 1991. NBC, which broadcast the game, said the actual audience was likely to have been higher because official ratings did not count people watching in locations such as sports bars. An additional 600,000 viewers watched the game with Spanish-language announcers on NBC's Universo cable network. NBC Sports also offered a live stream to computers and tablets - attracting an average 800,000 people, according to Adobe Analytics. After the game, an estimated 26.5 million people kept their TVs on NBC to watch an episode of James Spader's crime drama The Blacklist, making it the most-watched scripted programme on the network in more than a decade. Next year's Super Bowl will be played in Santa Clara, California, and will be broadcast by CBS. \newline \textbf{Ground Truth Summary: }An estimated average of 114.4 million people watched Sunday's Super Bowl on NBC, according to initial figures. } \\
  \hline
  Algorithm & High nli, Low rouge & Mid nli, Mid rouge & Low nli, High rouge \\
  \hline
  full-CLP & The Super Bowl has attracted a huge audience. & Some 120.8 million people watched the Super Bowl on TV on Sunday. & US viewers watched the Super Bowl in the United States with 120.8 million viewers. \\
  \hline
  attn-CLP & US viewers have watched the Super Bowl. & US viewers have watched the Super Bowl attracting more than 12 million people. & US viewers have watched the Super Bowl attracting more than 12 million people to watch the game. \\
  \hline
  logit-CLP & US viewers watched the Super Bowl with a huge audience. & US viewers watched the Super Bowl with a huge audience. & US viewers watched the Super Bowl with a huge audience. \\
  \hline
  prompting & Some 120.8 million viewers tuned in to the Super Bowl in the United States. & Some 120.8 million viewers tuned in to the Super Bowl in the United States. & Some 120.8 million viewers tuned in to the Super Bowl in the United States. \\
  \hline
\end{longtable}

\subsubsection{Document 16}
\begin{longtable}{|p{\sizeOfAlgorithmCol}|p{\sizeOfSummaryCol}|p{\sizeOfSummaryCol}|p{\sizeOfSummaryCol}|}
  \hline
  \multicolumn{4}{|p{\sizeOfLongTable}|}{\textbf{Document \#16}. summarize: Colombian leader Juan Manuel Santos and the Farc rebel commander known as Timochenko signed the deal in an emotional ceremony on Monday evening. "I would like to ask for forgiveness for all the pain that we may have caused during this war," he said. The guests at the ceremony in Cartagena cheered when Timochenko apologised. Some shouted "Yes, we can!" while Farc members and heads of state from Latin America rose to their feet on the stage and applauded. The ceremony which marks the end of 52 years of armed conflict was broadcast live and shown on giant screens in the capital, Bogota, and other large cities. Farc rebels gathered in a number of camps also followed the broadcast. There was so much symbolism in this historic signing - a pen made from a bullet to sign the peace deal, the singing of Beethoven's Ode to Joy, everyone dressed in white. President Santos said this historic moment was a message from Colombia to the world: no more war. "No more war," the crowd chanted in return. This was the first time Timochenko addressed the nation live on TV. He promised the Farc would give up its guns, and more than that, he asked for forgiveness. It earned him a standing ovation. That would have been unthinkable not long ago. But after 50 years of war, many Colombians still aren't ready to forgive. As President Santos put it, the hard work of building peace now lies ahead. Under the deal, the Farc will now become a political party. The rebels will take part in legislative elections in 2018, and - as part of the peace deal - they will be guaranteed a minimum of 10 seats in Congress for their first two legislative periods. President Santos addressed the Farc rebels directly and praised their decision to lay down arms. "Swapping bullets for votes and weapons for ideas is the bravest and most intelligent decision that any rebel group could take," he said. "When you begin your return to society (...) as head of state of the homeland that we all love, I welcome you to democracy." Timochenko said that the rebels would not take up arms anymore. "War is over, we're starting to build peace," he said before his speech was interrupted by the sound of Colombian Air Force fighter jets flying over the venue. Looking up startled, Timochenko's worried expression then turned into a smile and he joked: "Well, this time they came to pay their respects to peace and not to drop bombs." A number of Farc leaders, such Raul Reyes and Mono Jojoy, were killed in bombing raids over the past decade. The peace deal was reached after four years of formal peace talks in the Cuban capital, Havana, which were preceded by two years of secret talks. But before it can come into force it will have to be approved by Colombians in a public vote scheduled for Sunday. While recent polls suggest that a majority of Colombians back the deal, there has been vocal opposition to the agreement. Those in the "no" camp, which is led by former President Alvaro Uribe, say that they do not trust the Farc to stick to their promises. They are also angry that Farc leaders will be entering Congress without serving time in jail for crimes committed during the conflict. President Santos has stressed that there is "no Plan B". He said that if the Colombian people were to reject the peace deal, the country would face more conflict. 1964: Set up as armed wing of Communist Party 2002: At its height, it had an army of 20,000 fighters controlling up to a third of the country. Senator Ingrid Betancourt kidnapped and held for six years along with 14 other hostages 2008: The Farc suffers a series of defeats in its worst year 2012: Start of peace talks in Havana 2016: Definitive ceasefire Full timeline of Farc conflict \newline \textbf{Ground Truth Summary: }The leader of the Farc rebel group has apologised to the victims of Colombia's armed conflict which ended with Monday's signing of a peace deal. } \\
  \hline
  Algorithm & High nli, Low rouge & Mid nli, Mid rouge & Low nli, High rouge \\
  \hline
  full-CLP & A historic peace deal between Colombia's Farc rebels and the country's government has been signed. & Colombia has signed a peace deal with the Farc rebel group to end the 52-year conflict. & Colombia has signed a peace deal with the Farc rebel group to end the 50-year conflict. \\
  \hline
  attn-CLP & Colombia's Farc rebels have signed a peace deal. & Colombia's Farc rebel group have signed a peace deal to end the country's 52-year conflict. & Colombia's Farc rebel group have signed a peace deal to end the country's 52-year conflict. \\
  \hline
  logit-CLP & Colombia has signed a peace deal with the Farc rebels. & Colombia has signed a peace deal with the Farc rebels to end the 52-year war. & Colombia has signed a peace deal with the Farc rebels to end the 52-year war. \\
  \hline
  prompting & Colombia's Farc rebel group has signed a peace deal with the country. & Colombia's Farc rebel group has signed a peace deal with the country. & Colombia's Farc rebel group has signed a peace deal with the country. \\
  \hline
\end{longtable}

\newpage
\subsection{Generations for NLI v. TLDR}
In this section, we provide sample summaries from \clpall, \clpa, \clpl and the \clpp baseline on NLI v. TLDR where we evaluate three weightings $w_1=(0.8,0.2),w_2=(0.5,0.5),w_3=(0.2,0.8)$.

\subsubsection{Document 2}
\begin{longtable}{|p{\sizeOfAlgorithmCol}|p{\sizeOfSummaryCol}|p{\sizeOfSummaryCol}|p{\sizeOfSummaryCol}|}
  \hline
  \multicolumn{4}{|p{\sizeOfLongTable}|}{\textbf{Document \#2}. summarize: Two-time major winner McIlroy, 25, and Danish former world number one Wozniacki, 23, announced their engagement on New Year's Eve. Media playback is not supported on this device "The problem is mine," McIlroy said. "The wedding invitations issued at the weekend made me realise that I wasn't ready for all that marriage entails." The couple had been together for more than two years. McIlroy is currently at Wentworth for the BMW PGA Championship, the European Tour's flagship event which starts on Thursday. Wozniacki is set to compete in the French Open in Paris starting on 25 May. In a statement issued through his management company, McIlroy added: "There is no right way to end a relationship that has been so important to two people. "I wish Caroline all the happiness she deserves and thank her for the great times we have had." "The news has shocked the worlds of both golf and tennis, particularly with the timing of the split. McIlroy is due to start a crucial run of tournaments, while Wozniacki is playing the upcoming French Open." Speaking at Wentworth on Wednesday, McIlroy said the decision to end the engagement had been mutual and amicable. He said it was a difficult time, but that the move was best for both of them. McIlroy won the US Open in 2011 and US PGA Championship the following year. Wozniacki, who has reached the final of one grand slam tournament - the US Open in 2009 - is currently ranked 13th in the world. When McIlroy announced their engagement, he tweeted: "I have a feeling it is going to be a great year." \newline \textbf{Ground Truth Summary: }Northern Irish golf star Rory McIlroy has broken off his engagement with tennis player Caroline Wozniacki. } \\
  \hline
  Algorithm & High nli, Low tldr & Mid nli, Mid tldr & Low nli, High tldr \\
  \hline
  full-CLP & Rory McIlroy and Caroline Wozniacki have announced they are splitting up. & Rory McIlroy has announced that he and Caroline Wozniacki have split up after two years of passionate commitment. & Rory McIlroy has announced he has ended his engagement with tennis player Caroline Wozniacki, stressing the timing of the split was "shocking" and that the move was best for both partners. \\
  \hline
  attn-CLP & Rory McIlroy has announced that he is ending his engagement with former tennis partner Caroline Wozniacki. & Rory McIlroy has announced he has ended his engagement with tennis player Caroline Wozniacki, saying the split was mutual and mutually amicable. & Rory McIlroy has announced he has ended his engagement with tennis player Caroline Wozniacki, saying the split was mutual and mutually amicable. \\
  \hline
  logit-CLP & Rory McIlroy has announced he has ended his engagement with Caroline Wozniacki. & Rory McIlroy has announced he has ended his engagement with Caroline Wozniacki after realising he was not ready for the commitment. & Rory McIlroy has announced he has ended his engagement with Caroline Wozniacki after realising he was not ready for the commitment. \\
  \hline
  prompting & Rory McIlroy has announced he has ended his engagement with Caroline Wozniacki. & Rory McIlroy has announced he has ended his engagement with tennis player Caroline Wozniacki, saying the split is "no right way to end a relationship that has been so important to two people". & Rory McIlroy has announced he has ended his engagement with tennis player Caroline Wozniacki, saying the split is "no right way to end a relationship that has been so important to two people". \\
  \hline
\end{longtable}

\subsubsection{Document 3}
\begin{longtable}{|p{\sizeOfAlgorithmCol}|p{\sizeOfSummaryCol}|p{\sizeOfSummaryCol}|p{\sizeOfSummaryCol}|}
  \hline
  \multicolumn{4}{|p{\sizeOfLongTable}|}{\textbf{Document \#3}. summarize: The Battle of Britain Memorial Flight's Lancaster, known as Thumper, based at RAF Coningsby, took the skies for a test flight on Monday. The Lancaster, one of only two in the world able to fly, missed most of the 2015 display season. Squadron Leader Martin Morris said a schedule for subsequent flights will be announced over the next few weeks. Sqd Ldr Morris, who heads up the Battle of Britain Memorial Flight, said: "Spares and parts are difficult to find and some had to be manufactured. "The aluminium for the bulkhead had to be sourced from the same type of aluminium as the original aircraft - so it has not been without challenge - but our engineers have succeeded." Hundreds of people turned out to watch as Thumper took to the skies at about 14:00 BST. \newline \textbf{Ground Truth Summary: }The last airworthy Lancaster bomber in Britain has flown for the first time since being grounded by a fire in May. } \\
  \hline
  Algorithm & High nli, Low tldr & Mid nli, Mid tldr & Low nli, High tldr \\
  \hline
  full-CLP & A WWII aircraft that missed most of the plane display season has returned to the skies. & A WWII airworthy Lancaster aircraft which missed much of the 2015 display season has taken to the skies. & A WWII vintage Lancaster aircraft has successfully tested flight following considerable engineering and spares challenges, with a schedule for subsequent flights to be announced in the coming weeks. \\
  \hline
  attn-CLP & A WWII aircraft that missed most of the plane display season has successfully flown. & A WWII Lancaster that missed most of the plane display season has successfully flown for an initial test flight. & A WWII Lancaster that missed most of the plane display season has successfully flown for an initial test flight after overcoming "not without challenges" in sourcing spare parts and aluminium. \\
  \hline
  logit-CLP & A WWII airworthy Lancaster has flown for the first time since last year. & A WWII airworthy Lancaster has flown for the first time since last year after an extensive overhaul which proved challenging, an RAF unit has said. & A WWII airworthy Lancaster has performed its first test fly since it missed much of 2015 display season, with organisers promising to announce future flights in coming weeks. \\
  \hline
  prompting & A historic aircraft that missed most of the plane display season has successfully flown. & A historic Lancaster bomber has successfully tested for the first time after some challenging parts had to be manufactured. & A replica Lancaster aircraft which missed most of the 2015 display season has flown for an initial test flight after overcoming "difficult" requirements, including the need to source the exact aluminium type as the original. \\
  \hline
\end{longtable}

\subsubsection{Document 6}
\begin{longtable}{|p{\sizeOfAlgorithmCol}|p{\sizeOfSummaryCol}|p{\sizeOfSummaryCol}|p{\sizeOfSummaryCol}|}
  \hline
  \multicolumn{4}{|p{\sizeOfLongTable}|}{\textbf{Document \#6}. summarize: After the draft deal was published two weeks ago, an irate editorial in the French newspaper Le Monde fumed that the concessions made to Britain were "exorbitant". And while some in Mr Cameron's own Conservative Party dismiss his renegotiation as a watery fudge, politicians in Paris have a very different view. Of particular concern here is the question of future financial regulation. UK ministers, too, think this is the most important part of the whole renegotiation. The issue is this: The majority of EU countries that have the euro are integrating further and building a banking union with common rules and regulations. That process will deepen in the coming years. Mr Cameron wants legal guarantees that the eurozone will not rig the rules in a way that could harm the interests of countries that do not have the euro. But France frets that the UK is seeking special treatment for the City of London, that could harm the single market in financial services. France is also concerned a new mechanism that would allow non-euro countries to escalate their concerns could clog up decision-making. The UK insists it is not asking for any veto or competitive advantage, but this is likely to be the focus of the hastily arranged talks here this evening. President Hollande also faces an election next year - and while he wants the UK to stay in the EU, he will not do a deal that gives his opponents a chance to claim he sold out to Britain. EU renegotiation: Did Cameron get what he wanted? Referendum timeline: What will happen when? The view from Europe: What's in it for the others? More: BBC News EU referendum special \newline \textbf{Ground Truth Summary: }Prime Minister David Cameron is to meet French President Francois Hollande in Paris shortly - the latest leg in his drive to negotiate new EU membership terms for the UK. } \\
  \hline
  Algorithm & High nli, Low tldr & Mid nli, Mid tldr & Low nli, High tldr \\
  \hline
  full-CLP & Talks to re-build the EU with David Cameron are being held in Paris this evening. & As Britain prepares to begin negotiations on a deal with the European Union, debate is likely to turn to the future of financial regulation, and the way this could harm non-euro countries in the single market. & Britain's discussions with French President Hollande on a renegotiation of the EU's terms for the UK's bid to stay remain closely watched but France is concerned that the UK's request for special treatment in financial services may seriously harm the single market in Europe. \\
  \hline
  attn-CLP & Britain's discussions with French officials on a deal to resurrect Britain in the EU are likely to be marked by a number of concerns. & Britain's discussions with French officials on a deal to resurrect Britain in the EU have drawn some ire, with French politicians concerned about future financial regulation issues. & Britain's discussions with French officials on a deal to resurrect Britain in the EU this evening will likely be dominated by issues including future financial regulation and concerns that the UK is seeking special treatment for the City of London. \\
  \hline
  logit-CLP & Prime Minister David Cameron is negotiating with France on a renegotiated EU membership deal that will set out the UK's demands in terms of banking and financial regulation. & Prime Minister David Cameron's proposed EU renegotiation has faced heated criticism in France, with some worrying about the UK's bid to extract special treatment for London. & Prime Minister David Cameron's proposed EU renegotiation has faced heated criticism in France, with some worrying about the UK's bid to extract special treatment for London in the single market. \\
  \hline
  prompting & There is concern that recent EU negotiations on Britain's membership of the EU have left some countries concerned. & As Britain prepares to begin negotiations on a deal with the European Union, debate is about whether British ministers are getting what they want and whether France harbours fears that the UK is seeking special treatment for the City of London. & As Britain prepares to begin negotiations on a deal with the European Union, debate is about the crucial issue of future financial regulation - and with France worried over fears that the UK is seeking special treatment for City of London. \\
  \hline
\end{longtable}

\newpage
\section{Theory of Logit Mixing for Multi-Objective Finetuning}\label{app:theory}
Recall that \clpl (described in \cref{sec:parameter-conditioning-instantiations}) amounts to linear combination of the logits.
In this section, we show that mixing the logits of expert policies for each individual reward has provable guarantees under coverage conditions. We also provide a counterexample that rules out the optimality of zero-shot methods when this coverage condition is not satisfied. For simplicity, we focus on the two reward case as the general case follows naturally.

In this section, we will use $s$ to denote the input prompt, $a$ to denote the output generation and $|\Acal|$ to the cardinality of possible outputs.
Recall that the optimal policy for a fixed $(\alpha,R)$ has the form: $\pi^\star_{\alpha,R}(a\mid s)\propto \piref(a\mid s)\exp((1-\alpha)R(s,a)/\alpha)$ \citep{rafailov2023direct}.
Taking the logarithm gives
\begin{align*}
    \log\pi^\star_{\alpha,R}(a\mid s) = \log\piref(a\mid s)+\frac{1-\alpha}{\alpha}R(s,a)-\log Z_{\alpha,R}(s),
\end{align*}
where $Z_{\alpha,R}(s)=\sum_{a'}\piref(a\mid s)\exp((1-\alpha)R(s,a)/\alpha)$ in the partition function.
Fix any reward functions $R_1,R_2$ and weight $\lambda\in[0,1]$, and define $R_\lambda:= (1-\lambda) R_1+\lambda R_2$, we have
\begin{align*}
    \log\pi^\star_{\alpha,R_\lambda}(a\mid s)
    &= (1-\lambda)\log\pi^\star_{\alpha,R_1}(a\mid s) + \lambda\log\pi^\star_{\alpha,R_2}(a\mid s)
    \\&+ (1-\lambda)\log Z_{\alpha,R_1}(s)+\lambda\log Z_{\alpha,R_2}(s)- \log Z_{\alpha,R_\lambda}(s).
\end{align*}
Since the partition terms are independent of $a$, this implies that linearly interpolating the logits of $\pi^\star_{\alpha,R_1}$ and $\pi^\star_{\alpha,R_2}$ produces logits of the optimal policy for the combined reward. In other words, the optimal policy for the combined reward can be expressed as a multiplicative interpolation of the two expert policies. This was also alluded to in \citet[Appendix B]{liu2024decoding}.

However, since we never know the optimal policy in practice, we can only assume access to $\eps$-optimal policies for each individual reward. The following theorem quantities the sensitivity to $\eps$ for the logit mixing approach.
\begin{restatable}{theorem}{logitMixingMO}\label{thm:logit-mixing-mo}
Fix any $\alpha,\lambda\in[0,1]$.
Suppose $\wh\pi_1$ is an $\eps$-optimal policy for $R_1$, \ie, $V^\star_{\alpha,R_1}-V_{\alpha,R_1}(\wh\pi_1)\leq\eps$. Similarly assume $\wh\pi_2$ is $\eps$-optimal for $R_2$. Let $\wh\pi_\lambda$ be the logit mixing of $\wh\pi_1,\wh\pi_2$ as described above. Then,
\begin{align*}
    V_{\alpha,R_\lambda}^\star-V_{\alpha,R_\lambda}(\wh\pi_\lambda)\leq \eps\cdot\prns{ \exp(\eta^2/8)\prns{(1-\lambda)C_{\wh\pi_2,\wh\pi_1}^\lambda + \lambda C_{\wh\pi_1,\wh\pi_2}^{1-\lambda}} + 4|\Acal|p_{\min}^{-1} },
\end{align*}
where $\eta$ is the maximum $\ell_\infty$ logit value of $\wh\pi_1$ and $\wh\pi_2$, and $p_{\min}=\min_{i\in\{1,2\}}\min_{x,y}\wh\pi_i(y\mid x)$ is the minimum probability of an action.
\end{restatable}

\paragraph{Counterexample with bad coverage.}
In \cref{sec:counterexample-for-zero-shot}, we describe a counterexample for which the zero-shot logit mixing approach provably fails due to lack of coverage.
In the following experiment, we also find that zero-shot RS cannot learn the Pareto-optimal policy for this task, whereas \clp ultimately learns the Pareto optimal policy.
\begin{figure}[!h]
    \centering
    \includegraphics[width=0.5\linewidth]{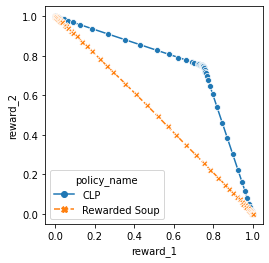}
    \caption{Pareto-fronts of RS and \clp in the counterexample of \cref{sec:counterexample-for-zero-shot}.}
    \label{fig:counterexample-zero-shot}
\end{figure}

\subsection{Proofs}

\begin{proof}[Proof of \cref{thm:logit-mixing-mo}.]
By \cref{lem:kl-regularized-regret}, we can focus on bounding $\KL(\wh\pi_\lambda(s)\Mid\pi^\star_\lambda(s))$, we can be simplified as follows:
\begin{align*}
    &\KL(\wh\pi_\lambda(s)\Mid\pi^\star_\lambda(s))
    \\&=\EE_{a\sim\wh\pi_\lambda}[ \log\wh\pi_\lambda(a\mid s)- \log\pi^\star_\lambda(a\mid s)]
    \\&=\EE_{a\sim\wh\pi_\lambda}[ (1-\lambda)(\log\wh\pi_1(a\mid s)-\log\pi^\star_1(a\mid s)) + \lambda(\log\wh\pi_2(a\mid s)-\log\pi^\star_2(a\mid s))]
    \\&+ (1-\lambda)(\log\wh Z_1(s)-\log Z^\star_1(s))+\lambda(\log\wh Z_2(s)-\log Z^\star_2(s)) - (\log\wh Z_\lambda(s)-\log Z^\star_\lambda(s)).
\end{align*}

\paragraph{Bounding the log partition terms.}
First, we bound the difference of log partition functions.
Let us parameterize the policies $\wh\pi_1(\cdot\mid s) = \sigma(\wh\theta_1(s))$ where $\wh\theta_1(s)\in\RR^{A}$ (where $A=|\Acal|$) and $\sigma$ is the softmax operation, \ie, $\sigma(x)[i] = \frac{\exp(x_i)}{\sum_j \exp(x_j)}$.
Without loss of generality, we fix $\wh\theta_1(s)[A] = 0$ and thus can rewrite $\wh Z_1(s)=1+\sum_{a=1}^{A-1}\exp(\wh\theta(s)[a])$.
In particular, we have $\log\wh\pi_1(a\mid s) = \theta(s)[a]-\log\wh Z_1(s)$.

We abuse notation and let $Z(\theta(s)) = \sum_a\exp(\theta(s)[a])$.
Notice that $\KL(\wh\pi(s)\Mid\pi^\star(s))$ is the Bregman divergence of $f(\theta(s)):=\log Z(\theta(s))$ because $\nabla_\theta f(\theta(s)) = \sigma(\theta(s))=\pi_\theta(s)$, \ie, we have,
\begin{align*}
    \KL(\wh\pi(s)\Mid\pi^\star(s)) = \EE_{a\sim\wh\pi(s)}[ \wh\theta(a\mid s)-\theta^\star(a\mid s) ] + \log Z(\theta^\star(s)) - \log Z(\wh\theta(s)) = \Delta_f(\theta^\star(s),\wh\theta(s)).
\end{align*}

By \cref{lem:strong-convexity}, we have that $f$ is $\sigma$-strongly convex where $\sigma=\frac{p_{\min}}{A}$, and hence:
\begin{align*}
    \|\theta^\star_1(s)-\wh\theta_1(s)\|_2\leq \frac{2}{\sigma}\Delta_f(\theta^\star_1(s),\wh\theta_1(s)) = \frac{2}{\sigma}\KL(\wh\pi_1(s)\Mid\pi^\star_1(s)) \leq 2\eps\sigma^{-1},
\end{align*}
where the last inequality is due to the premise.

Since $\nabla f(\theta) = \sigma(\theta)$, its $\ell_2$-norm is bounded by $1$ and thus $f$ is $1$-Lipschitz. Hence:
\begin{align*}
    \abs{f(\theta^\star_1(s))-f(\wh\theta_1(s))}\leq \|\theta^\star_1(s)-\wh\theta_1(s)\|_2 \leq 2\eps\sigma^{-1}.
\end{align*}
Thus, the log partition terms are bounded as follows:
\begin{align*}
    &(1-\lambda)(\log\wh Z_1(s)-\log Z^\star_1(s))+\lambda(\log\wh Z_2(s)-\log Z^\star_2(s)) - (\log\wh Z_\lambda(s)-\log Z^\star_\lambda(s))
    \\&\leq (1-\lambda)2\eps\sigma^{-1}+\lambda 2\eps\sigma^{-1} + \|\theta^\star_\lambda(s)-\wh\theta_\lambda(s)\|_2
    \\&\leq 2\eps\sigma^{-1} + (1-\lambda)\|\theta^\star_1(s)-\wh\theta_1(s)\|_2 + \lambda\|\theta^\star_2(s)-\wh\theta_2(s)\|_2
    \\&\leq 4\eps\sigma^{-1}.
\end{align*}

\paragraph{Bounding the on-policy term.}
To perform a change of measure, we can compute the density of $\wh\pi_\lambda$ with $\wh\pi_1$ and $\wh\pi_2$ as follows:
\begin{align*}
    \textstyle\frac{\wh\pi_\lambda(a\mid s)}{\wh\pi_1(a\mid s)}
    &\textstyle= \prns{ \frac{\wh\pi_2(a\mid s)}{\wh\pi_1(a\mid s)} }^\lambda \frac{\wh Z_1(s)^{1-\lambda}\wh Z_2(s)^\lambda}{\wh Z_\lambda(s)},
    \\\textstyle\frac{\wh\pi_\lambda(a\mid s)}{\wh\pi_2(a\mid s)}
    &\textstyle= \prns{ \frac{\wh\pi_1(a\mid s)}{\wh\pi_2(a\mid s)} }^{1-\lambda} \frac{\wh Z_1(s)^{1-\lambda}\wh Z_2(s)^\lambda}{\wh Z_\lambda(s)}.
\end{align*}
Thus,
\begin{align*}
    &\EE_{a\sim\wh\pi_\lambda}[ (1-\lambda)(\log\wh\pi_1(a\mid s)-\log\pi^\star_1(a\mid s)) + \lambda(\log\wh\pi_2(a\mid s)-\log\pi^\star_2(a\mid s))]
    \\&\textstyle= \frac{\wh Z_1(s)^{1-\lambda}\wh Z_2(s)^\lambda}{\wh Z_\lambda(s)} \bigg( (1-\lambda) \EE_{a\sim\wh\pi_1(s)} \prns{ \frac{\wh\pi_2(a\mid s)}{\wh\pi_1(a\mid s)} }^\lambda (\log\wh\pi_1(a\mid s)-\log\pi^\star_1(a\mid s))
    \\&\textstyle\qquad\qquad\qquad\qquad\qquad + \lambda\EE_{a\sim\wh\pi_2(s)}\prns{ \frac{\wh\pi_1(a\mid s)}{\wh\pi_2(a\mid s)} }^{1-\lambda}(\log\wh\pi_2(a\mid s)-\log\pi^\star_2(a\mid s)) \bigg)
    \\&\textstyle\leq \exp(\eta^2/8) \prns{ (1-\lambda)C_{\wh\pi_2,\wh\pi_1}^\lambda + \lambda C_{\wh\pi_1,\wh\pi_2}^{1-\lambda} } \eps,
\end{align*}
where we used \cref{lem:ratio-of-softmax} to bound the factor in front, bounded the density ratios, and used the fact that $\KL(\wh\pi_1(s)\Mid\pi^\star_1(s))\leq\eps$ and $\KL(\wh\pi_2(s)\Mid\pi^\star_2(s))\leq\eps$ by premise.
\end{proof}

\begin{lemma}\label{lem:kl-regularized-regret}
For any $\pi$:
\begin{align*}
    V_{\alpha,R}(\pi) = \alpha\EE_{s\sim\mu}[ \log Z_{\alpha,R}(s) - \KL(\pi(s)\Mid\pi^\star_{\alpha,R}(s)) ].
\end{align*}
Therefore,
\begin{align*}
    V^\star_{\alpha,R}-V_{\alpha,R}(\pi) = \alpha\EE_{s\sim\mu}[\KL(\pi(s)\Mid\pi^\star_{\alpha,R}(s))].
\end{align*}
\end{lemma}
\begin{proof}
Recall that $\pi^\star_{\alpha,R}(a\mid s) = \frac{\piref(a\mid s)\exp((1-\alpha)R(s,a)/\alpha)}{Z_{\alpha,R}(s)}$, where $Z_{\alpha,R}(s) = \sum_{a'}\piref(a'\mid s)\exp((1-\alpha)R(s,a')/\alpha)$ is the partition function. Then,
\begin{align*}
    V_{\alpha,R}(\pi)
    &= \EE_{s\sim\mu,a\sim\pi(s)}[ (1-\alpha)R(s,a)-\alpha(\log\pi(a\mid s)-\log\piref(a\mid s)) ]
    \\&= \alpha\EE_{s\sim\mu}[ \log\pi^\star_{\alpha,R}(a\mid s)-\log\pi(a\mid s)+\log Z_{\alpha,R}(s) ]
    \\&= \alpha\EE_{s\sim\mu}[ \log Z_{\alpha,R}(s) - \KL(\pi(s)\Mid\pi^\star_{\alpha,R}(s)) ].
\end{align*}
\end{proof}

\begin{lemma}[Hoeffding's Lemma]\label{lem:hoeffding}
Let $X$ be a random variable such that $X\in[a,b]$ w.p. $1$. Then for all $\lambda\in\RR$,
\begin{align*}
    \textstyle\log\EE\exp(\lambda X) \leq \lambda\EE X + \frac{\lambda^2(b-a)^2}{8}.
\end{align*}
\end{lemma}

\begin{lemma}\label{lem:ratio-of-softmax}
For any $x,y\in\RR^n$ and $\lambda\in[0,1]$, we have
\begin{align*}
    \log\prns{ \frac{(\sum_i\exp(x_i))^{1-\lambda}(\sum_i\exp(y_i))^\lambda}{\sum_i\exp((1-\lambda)x_i+\lambda y_i)} } \leq B^2/8,
\end{align*}
where $B = \|x\|_\infty\vee\|y\|_\infty$.
\end{lemma}
\begin{proof}
\begin{align*}
    &\textstyle(1-\lambda)\log\sum_i\exp(x_i)+\lambda\log\sum_i\exp(y_i)
    \\&=(1-\lambda)\log\EE_{i\sim\Ucal([n])}\exp(x_i)+\lambda\log\EE_{i\sim\Ucal([n])}\exp(y_i) + \log(n)
    \\&\leq (1-\lambda)\EE_{i\sim\Ucal([n])}x_i+(1-\lambda)B^2/8 + \lambda\EE_{i\sim\Ucal([n])}y_i+\lambda B^2/8 + \log(n) \tag{\cref{lem:hoeffding}}
    \\&= \EE_{i\sim\Ucal([n])}[(1-\lambda)x_i+\lambda y_i] + \log(n) + B^2/8
    \\&\leq \log\EE_{i\sim\Ucal([n])}\exp((1-\lambda)x_i+\lambda y_i)+\log(n)+B^2/8 \tag{Jensen's inequality}
    \\&\textstyle= \log\sum_i\exp((1-\lambda)x_i+\lambda y_i)+B^2/8.
\end{align*}
\end{proof}

\begin{lemma}\label{lem:strong-convexity}
Let $x\in\RR^{n-1}$ and $f(x)=\log(1+\sum_i\exp(x_i))$. Then $f$ is $\sigma$-strongly convex for $\sigma\geq\frac{p_{\min}}{n}$ where $p_{\min}=\min(\frac{1}{1+\sum_i\exp(x_i)}, \frac{\exp(x_1)}{1+\sum_i\exp(x_i)}, \dots, \frac{\exp(x_{n-1})}{1+\sum_i\exp(x_i)}).$
\end{lemma}
\begin{proof}
The gradient and Hessian of $f$ are:
\begin{align*}
    &\nabla f(x)=\sigma(x)
    \\&\nabla^2 f(x) = \op{diag}(\sigma(x))-\sigma(x)\sigma(x)^\top.
\end{align*}
Now, apply \cref{lem:strong-convexity-two} to the Hessian, which completes the proof.
\end{proof}
\begin{lemma}\label{lem:strong-convexity-two}
Let $p\in\RR^{n-1}_+$ such that $\sum_ip_i < 1$. Then $H = \op{diag}(p)-pp^\top$ satisfies $\lambda_{\min}(H)\geq \frac{p_{\min}}{n}$ where $p_{\min}=\min(p_1,p_2,\dots,p_{n-1},1-\sum_ip_i)$.
\end{lemma}
\begin{proof}
To lower bound the minimum eigenvalue, we'll lower bound the quadratic form for any $v\in\RR^{n-1}$ with $\|v\|_2=1$. Let $\mu=\sum_ip_iv_i$ and note that
\begin{align*}
    \textstyle\sum_ip_i(v_i-\mu)^2 = \sum_ip_iv_i^2-\mu^2 + (\sum_ip_i - 1)\mu^2.
\end{align*}
Hence,
\begin{align*}
    v^\top Hv
    &\textstyle= \sum_i p_iv_i^2 - \mu^2
    \\&\textstyle=\sum_i p_i(v_i-\mu)^2 + (1-\sum_i p_i)\mu^2
    \\&\textstyle\geq p_{\min}(\sum_i(v_i-\mu)^2 + \mu^2).
\end{align*}
It remains to show that $f(\mu) := \sum_i(v_i-\mu)^2 + \mu^2\geq \frac{1}{n}$. Note that $f'(\mu) = 2n\mu - 2\sum_i v_i$ and $f''(\mu) = 2n$. Thus, $f$ is convex and its minimizer is $\wt\mu := \frac{1}{n}\sum_i v_i$. Finally,
\begin{align*}
    f(\wt\mu)
    &\textstyle= n\wt\mu^2 + \sum_i v_i^2 - 2\sum_i v_i\wt\mu
    \\&\textstyle= 1+\wt\mu(n\wt\mu-2\sum_i v_i) \tag{$\|v\|_2=1$}
    \\&\textstyle= 1- \frac{1}{n}(\sum_iv_i)^2
    \\&\textstyle\geq 1-\frac{n-1}{n} = \frac{1}{n}. \tag{ $(\sum_iv_i)^2\leq (n-1)\sum_i v_i^2 = n-1$ }
\end{align*}
Thus, we've shown that $v^\top H v \geq \frac{p_{\min}}{n}$ for all $v\in\RR^{n-1}$ with $\|v\|_2=1$, which implies the claim.
\end{proof}

\end{document}